%% file: main.tex
\documentclass[runningheads]{llncs}

\usepackage{eccv}

\usepackage{eccvabbrv}

\usepackage{graphicx}
\usepackage{booktabs}
\usepackage[accsupp]{axessibility}  %
\usepackage{bm}
\usepackage{tikz}
\definecolor{darkpastelgreen}{rgb}{0.01, 0.75, 0.24}
\definecolor{darkgreen}{rgb}{0.00, 0.8, 0.2}
\definecolor{darkyellow}{rgb}{0.96, 0.75, 0.00}
\definecolor{custompurple}{RGB}{128, 0, 128}

\usepackage[breaklinks,colorlinks,citecolor=eccvblue]{hyperref}

\definecolor{darkpastelgreen}{rgb}{0.01, 0.75, 0.24}
\definecolor{darkgreen}{rgb}{0.00, 0.8, 0.2}
\definecolor{darkyellow}{rgb}{0.96, 0.75, 0.00}
\definecolor{historyblue}{rgb}{0.36,0.6,0.78}

\begin{document}

\title{Accelerating Online Mapping and Behavior Prediction via Direct BEV Feature Attention} 

\titlerunning{Accelerating Online Mapping and Behavior Prediction}

\author{Xunjiang Gu\inst{1} \and
Guanyu Song\inst{1} \and
Igor Gilitschenski\inst{1,2} \and
Marco Pavone\inst{3,4} \and
Boris Ivanovic\inst{3}}

\institute{$^1$University of Toronto \hspace{0.1cm} $^2$Vector Institute \hspace{0.1cm} $^3$NVIDIA Research \hspace{0.1cm} $^4$Stanford University\\
\vspace{0.1cm}\email{\{alfred.gu, guanyu.song\}@mail.utoronto.ca, gilitschenski@cs.toronto.edu, \{mpavone, bivanovic\}@nvidia.com, pavone@stanford.edu}}

\authorrunning{X.~Gu et al.}

\maketitle
\input{sec/0_abstract}    
\input{sec/1_introduction}
\input{sec/2_related_work}

\input{sec/3_methods}
\input{sec/4_experiments}
\input{sec/5_conclusion}

\bibliographystyle{splncs04}
\bibliography{main}

\input{sec/6_suppl}
\end{document}

%% file: sec/0_abstract.tex
\begin{abstract}
\setcounter{footnote}{0}
\vspace{-0.3cm}
Understanding road geometry is a critical component of the autonomous vehicle (AV) stack. While high-definition (HD) maps can readily provide such information, they suffer from high labeling and maintenance costs. Accordingly, many recent works have proposed methods for estimating HD maps online from sensor data. The vast majority of recent approaches encode multi-camera observations into an intermediate representation, e.g., a bird's eye view (BEV) grid, and produce vector map elements via a decoder. While this architecture is performant, it decimates much of the information encoded in the intermediate representation, preventing downstream tasks (e.g., behavior prediction) from leveraging them. In this work, we propose exposing the rich internal features of online map estimation methods and show how they enable more tightly integrating online mapping with trajectory forecasting\footnote{Code: \url{https://github.com/alfredgu001324/MapBEVPrediction}}. In doing so, we find that directly accessing internal BEV features yields up to \textbf{73\%} faster inference speeds and up to \textbf{29\%} more accurate predictions on the real-world nuScenes dataset.
\keywords{Autonomous Driving \and Online HD Map Estimation \and Behavior Prediction}
\end{abstract}

%% file: sec/1_introduction.tex
\section{Introduction}
\label{sec:intro}

Perceiving the static environment surrounding an autonomous vehicle (AV) is a critical task for autonomous driving, providing geometric information (e.g., road layout) to downstream behavior prediction and motion planning modules. Traditionally, high-definition (HD) maps have served as the backbone for this understanding, offering centimeter-level geometries for road boundaries, lane dividers, lane centerlines, pedestrian crosswalks, traffic signs, road markings, and more. They have proven to be an indispensable part of enhancing AV situational awareness and navigational judgment in downstream prediction tasks. However, despite their undeniable utility, collecting and maintaining HD maps is labor-intensive and costly, which limits their scalability. 

\begin{figure}[t]
    \centering
    \includegraphics[width=\linewidth]{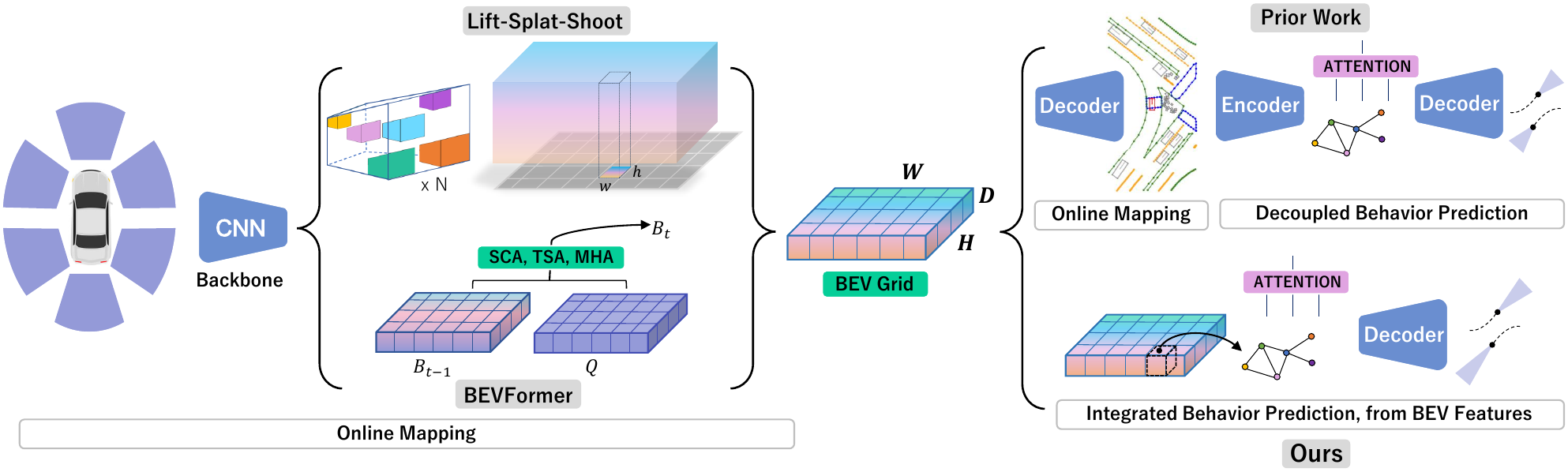}
    
    \vspace{-0.3cm}
    
    \caption{Online map estimation approaches predominantly encode multi-camera observations into a canonical BEV feature grid prior to decoding vectorized map elements. In this work, we propose deeply integrating online mapping with downstream tasks through direct access to the rich BEV features of online map estimation methods.}
    \label{fig:arch}

    \vspace{-0.6cm}
    
\end{figure}

In recent years, online HD map estimation methods have emerged as an alternative, aiming to predict HD map information directly from sensor observations. Starting from (multi-)camera images and optionally LiDAR pointclouds, state-of-the-art HD map estimation methods broadly employ an encoder-decoder neural network architecture (\cref{fig:arch}). An encoder first converts the sensor observations to a bird's eye view (BEV) grid of features. A decoder then predicts the location and semantic type of map elements from the BEV features. The resulting road geometries are commonly structured as combinations of polylines and polygons per map element type (e.g., road boundaries, lane dividers, pedestrian crosswalks). Such online-estimated maps serve as a practical substitute for offline HD mapping, providing necessary scene context for downstream tasks such as behavior prediction and motion planning. As an example, recent work~\cite{GuSongEtAl2024} has shown success in coupling various map estimation methods with existing prediction frameworks, highlighting their potential to expedite the development of end-to-end AV stacks.

While such encoder-decoder approaches produce accurate HD maps, as we will show in \cref{sec:expt}, the attention mechanisms employed in decoding are computationally expensive (occupying the majority of model runtime) and do not produce outputs with associated uncertainty, which limits the ability of downstream modules to account for uncertainty. Moreover, such an architecture prevents downstream tasks from leveraging the rich intermediate features generated in the encoder's perspective-view-to-bird's-eye-view (PV2BEV) transformation, decimating information that cannot be described by point sets. 

\textbf{Contributions.} Towards this end, we introduce three novel scene encoding strategies that leverage internal BEV features to improve the performance and accelerate the runtime of combined online mapping and behavior prediction systems. By directly leveraging BEV features, our proposed methods provide tighter integrations between map estimation and behavior prediction frameworks, achieving up to \textbf{73\%} faster system inference speeds and an up to \textbf{29\%} increase in downstream prediction accuracy on the real-world nuScenes dataset.

%% file: sec/2_related_work.tex
\section{Related Work}
\label{sec:related_work}

\subsection{Online Map Estimation}
Online map estimation methods focus on generating a representation of the static environment surrounding an autonomous vehicle from its sensor data. Initial approaches used 2D BEV rasterized semantic segmentations as world representations. These maps were produced by either transforming observations to 3D and collapsing along the $Z$-axis~\cite{philion2020lss,liu2022bevfusion} or by utilizing cross-attention in geometry-aware Transformer~\cite{VaswaniShazeerEtAl2017} models~\cite{can2021stsu,li2022bevformer}.

Recently, there has been a growth in vectorized map estimation methods that extend traditional BEV rasterization approaches. These methods employ an encoder-decoder architecture which regresses and classifies map elements in the form of polylines, polygons, and other curve representations~\cite{qiao2023bemapnet}. Initial methods such as SuperFusion~\cite{dong2022SuperFusion} and HDMapNet~\cite{li2022hdmapnet} combined both LiDAR point clouds and RGB images into a common BEV feature frame, with a subsequent hand-crafted post-processing stage to generate polyline map elements. To eliminate this post-processsing step, VectorMapNet~\cite{liu2022vectormapnet} and InstaGraM~\cite{shin2023instagram} propose end-to-end models for vectorized HD map estimation.

In parallel, HD map estimation has also been formulated as a point set prediction task in the MapTR line of work~\cite{MapTR, maptrv2} and its extensions~\cite{xu2023insightmapper}, yielding significant advancements in map estimation performance. To enable online inference from streaming observations, StreamMapNet~\cite{yuan2024streammapnet} introduces a memory buffer that incorporates temporal data from prior timesteps. As many of these methods are commonly employed today, in this work we show how BEV features from multiple diverse mapping approaches can be leveraged to improve integrated system performance.

\subsection{Map-Informed Trajectory Prediction}
Learning-based trajectory prediction approaches initially leveraged rasterized maps for semantic scene context~\cite{RudenkoPalmieriEtAl2019}. The rasterized map is treated as a top-down image and encoded via a convolutional neural network (CNN), concatenated with other scene context information (e.g., agent state history), and passed through the rest of the model~\cite{SalzmannIvanovicEtAl2020,Phan-MinhGrigoreEtAl2020,YuanWengEtAl2021,GillesSabatiniEtAl2021,IvanovicHarrisonEtAl2023}.
Recently, state-of-the-art trajectory prediction methods have increasingly shifted to directly encoding raw polyline information from vectorized HD maps, demonstrating significant improvements in prediction accuracy. Initial approaches~\cite{gao2020vectornet,liang2020lanegcn,ZhaoGaoEtAl2020,GillesSabatiniEtAl2022a,GillesSabatiniEtAl2022b} utilized graph neural networks (GNNs) to encode lane polylines as nodes and their interactions with agent trajectories as edges. Extending this insight, Transformer~\cite{VaswaniShazeerEtAl2017} architectures with attention over map and agent embeddings have been widely adopted by current state-of-art methods~\cite{liu2021mmtransformer,GuSunEtAl2021,zhou2022hivt,deo2023pgp}.

One recent related work investigates the integration between different combinations of map estimation and trajectory prediction models~\cite{GuSongEtAl2024}. In it, they propose exposing uncertainties from map element regression and classification to downstream behavior prediction.
In contrast to~\cite{GuSongEtAl2024}, our work focuses on exposing information from an earlier stage of online mapping (immediately following observation encoding). As we will show in~\cref{sec:expt}, our approach not only outperforms~\cite{GuSongEtAl2024}, it is also much more computationally efficient.

\subsection{End-to-End Driving Architectures}
End-to-end architectures present a promising approach for creating integrated stacks that can internally leverage more information, e.g., uncertainty, from upstream components. Recent works such as UniAD~\cite{hu2023uniad}, VAD~\cite{jiang2023vad}, and OccNet~\cite{tong2023occnet} demonstrate the feasibility and performance of incorporating both rasterized and vectorized HD map estimation within end-to-end driving. For example, UniAD~\cite{hu2023uniad} and OccNet~\cite{tong2023occnet} formulate online mapping as a dense prediction task, aiming to generate the semantics of map elements at a per-pixel or voxel granularity, whereas VAD focuses on producing vectorized HD map representations. In these architectures, the utility of mapping is twofold: it is both an auxiliary training task and an internal static world representation that aids downstream tasks. While these approaches lead to highly-integrated autonomy stacks, the use of rasterized or vectorized representations (rather than BEV features) leads to information loss and extra computational burden. Accordingly, our work is complementary in that our proposed strategies can be incorporated within end-to-end stacks to improve inference speeds as well as downstream prediction and planning accuracy.

%% file: sec/3_methods.tex
\section{Leveraging Online Mapping Features in Trajectory Prediction}
\label{sec:methods}
As mentioned in \cref{sec:related_work}, the majority of state-of-the-art online vectorized map estimation models adopt a BEV grid internally to featurize the surrounding environment in a geometry-preserving fashion. 
Our method focuses on leveraging these internal BEV representations by directly accessing them in trajectory prediction. In doing so, we improve the flow of information from mapping to prediction and can even accelerate the combined system's runtime by skipping map decoding altogether (depending on the predictor's need for lane information).

\textbf{Encoding Observations:} Feature extractors in map estimation models aim to transform inputs from various vehicle-mounted sensors (e.g., cameras and LiDAR) into a unified feature space. Note that our work focuses on multi-camera observations, in line with the majority of state-of-the-art map estimation approaches.
Formally, given a set of multi-view images \(I_t = \{I_1, \ldots, I_K\}\) at time $t$, map estimation models encode them using a standard backbone (e.g., ResNet-50~\cite{HeZhangEtAl2016}) to generate corresponding multi-view feature maps \(F_t = \{F_1, \ldots, F_K\}\). The 2D image features \(F_t\) are then converted into BEV features $B_t$ using a PV2BEV transformation. The two most common PV2BEV approaches are based on BEVFormer~\cite{li2022bevformer} and Lift-Splat-Shoot (LSS)~\cite{philion2020lss}. 

BEVFormer~\cite{li2022bevformer} is a Transformer-based architecture that converts multi-camera image features into BEV features. It employs a standard Transformer encoder with specific enhancements: BEV queries \(Q \in \mathbb{R}^{H \times W \times D}\), spatial cross-attention, and temporal self-attention. First, temporal information is queried from historical BEV features \(B_{t-1}\) through temporal self-attention,
\begin{equation}
\text{TSA}(Q_p, \{Q, B_{t-1}\}) = \sum_{V \in \{Q,B_{t-1}\}} \text{DeformAttn}(Q_p, p, V),
\end{equation}
where $Q_p \in \mathbb{R}^D$ is the query for a single BEV grid point $p = (h, w)$. The queries \(Q\) are then employed to gather spatial information from the multi-camera features \(F_t\) via a spatial cross-attention mechanism,
\begin{equation}
\text{SCA}(Q_p, F_t) = \frac{1}{|\mathcal{V}_{\text{hit}}|} \sum_{i \in \mathcal{V}_{\text{hit}}} \sum_{j=1}^{N_{\text{ref}}} \text{DeformAttn}(Q_p, \mathcal{P}(p, i, j), F_{i,t}),
\end{equation}
where $\mathcal{V}_{\text{hit}}$ denotes the camera views that contain $p$, $\mathcal{P}$ is the camera projection function from 3D world coordinates ($h, w$, and discrete height index $j$) to the 2D image plane of the $i^\text{th}$ camera.
This combined approach enables BEVFormer to efficiently understand temporal and spatial context, producing enhanced BEV features. As we will show in \cref{sec:abalation}, incorporating temporal information in BEV features is quite beneficial for trajectory prediction. 

Another common PV2BEV method is LSS~\cite{philion2020lss}. Its first stage (Lift) featurizes individual images and converts them into a shared 3D frame via ``unprojection", assigning multiple discrete depth points \((h, w, d) \in \mathbb{R}^3\) to each pixel in an image based on camera extrinsics and intrinsics. This forms a large point cloud with a 3D point at each depth per ray ($HWD$ points). The second stage (Splat) aggregates these points into a common BEV feature grid using an efficient pillar pooling method.

\textbf{Decoding Map Elements:} To produce vectorized map elements, most map prediction models employ a Transformer-based decoder. They broadly consist of a hierarchical query embedding mechanism alongside Multihead Self Attention and Deformable Attention to accurately predict complex, irregular map elements from BEV features. Instance and point-level queries are combined for dynamic feature interaction, followed by classification and regression heads that predict the type and location of map element vertices, respectively.
While such decoding architectures produce accurate maps, they are computationally expensive, and decoding occupies much of overall model runtime. MapTRv2~\cite{maptrv2} attempts to address this by introducing more streamlined attention mechanisms in its decoder. In StreamMapNet~\cite{yuan2024streammapnet}, a Multi-Point Attention mechanism is utilized alongside a streaming approach that preserves previous queries and BEV features, aiming to improve map estimation performance by incorporating temporal information. %

\textbf{Behavior Prediction Models:} Most state-of-the-art trajectory prediction models also leverage an encoder-decoder framework~\cite{RudenkoPalmieriEtAl2019}. The encoder is responsible for capturing the scene's context, such as vectorized map elements (e.g., road edges and centerlines) as well as agent trajectories. The decoder then utilizes these encoded representations to forecast the future motion of agents in the scene. In the encoder, vectorized map elements are commonly encoded as either nodes in a Graph Neural Network (GNN) or as tokens in a Transformer~\cite{VaswaniShazeerEtAl2017}. Two representative instantiations are DenseTNT~\cite{GuSunEtAl2021} and HiVT~\cite{zhou2022hivt}, respectively. 

At a high level, DenseTNT~\cite{GuSunEtAl2021} leverages the VectorNet~\cite{gao2020vectornet} hierarchical GNN context encoder to extract features from vectorized map elements. Agent trajectories and map element segments are first modeled as polyline subgraphs. Then, each resulting subgraph is further encoded as a node in a global GNN to capture their interactions.
On the other hand, the Transformer-based HiVT~\cite{zhou2022hivt} treats vectorized elements as sequences of tokens. Its hierarchical encoder consists of two stages: information within a local spatial window is encoded for each agent, followed by a global interaction encoder to model long-range interactions between agents.

Recent work~\cite{GuSongEtAl2024} has explored strategies for coupling online-estimated vectorized maps and the above prediction models. However, as we will show in \cref{sec:expt}, their prediction performance and computational runtime can be further improved by harnessing the BEV features present within online map estimation models. Directly using BEV features provides prediction models access to richer information than the original decoded sets of polylines and polygons.

In the remainder of this section, we outline three different strategies for incorporating BEV features in downstream behavior prediction. 

\subsection{Modeling Agent-Lane Interactions via BEV Feature Attention}
\label{sec:replace_attend}

Inspired by the approach taken in Vision Transformer (ViT)~\cite{dosovitskiy2020vit}, we treat map BEV features as an image, albeit with a channel dimension equal to the embedding dimension. We process the BEV tensor by first dividing it into a sequence of flattened BEV patches, i.e., a $N \times P^2 D$ tensor, where $D$ denotes the embedding dimension, $(P, P)$ represents the resolution of each image patch, and $N = HW / P^2$ is the total number of BEV patches. The flattened BEV features are then passed through a trainable linear projection to obtain an $N \times D$ patch embedding for each scene. This serves as the equivalent of an $N$-length input sequence for a Transformer, allowing for attention mechanisms to be applied. 

\begin{figure}[t]
    \centering
    \includegraphics[width=0.9\linewidth]{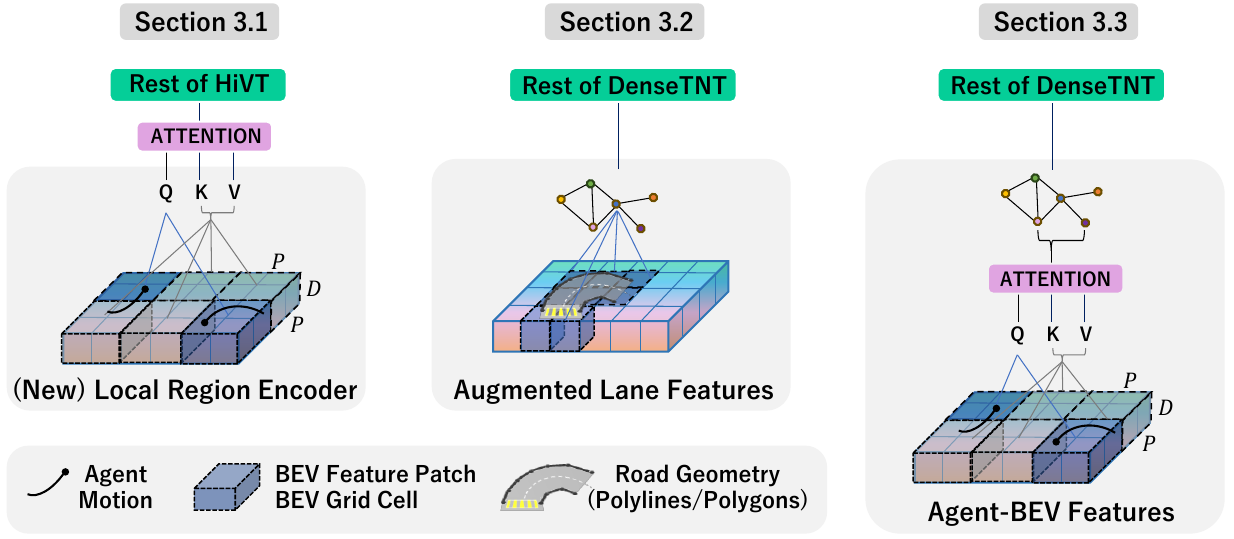}

    \caption{
    Three different strategies for incorporating BEV features in behavior prediction. \textbf{Left:} local region attention to encode agent-map interaction; \textbf{Middle:} augmenting lane vertices with BEV features; \textbf{Right:} replacing agent trajectories with temporal BEV features.}
    \label{fig:methods}

    \vspace{-0.5cm}
    
\end{figure}

In this first strategy for incorporating BEV features in behavior prediction, visualized in \cref{fig:methods}, we alter the modeling of agent-lane interactions. We select BEV grid patches that correspond to agent positions and attend to every other patch in the scene, providing an understanding of the environment surrounding the vehicle. Formally, 
\begin{equation}
\label{eqn:agent_bev_attn}
\mathbf{e}_A = \text{MHA}(Q_A, K_M, V_M),
\end{equation}
where MHA denotes multi-headed attention, $Q_A \in \mathbb{R}^{M \times D}$ denotes agent patches (queries), $K_M = V_M \in \mathbb{R}^{N \times D}$ denote all map patches (keys and values), and $\mathbf{e}_A$ are the resulting agent-BEV embeddings.
As an additional benefit, by computing attention with agent patches, the computational complexity is only $M \times N$ operations, where $M$ denotes the number of agents (most commonly, $M \ll N$). 

To verify our approach, we modify the state-of-the-art Transformer-based prediction model HiVT~\cite{zhou2022hivt}.
It employs a hierarchical Transformer structure with a low-level Transformer that encodes agent-lane interactions within a local neighborhood, followed by a high-level Transformer that models global interactions across the entire scene.
Note that, by replacing the local agent-lane interaction encoder with the agent-BEV attention in \cref{eqn:agent_bev_attn}, we completely remove the use of vectorized lane information in HiVT. The agent-BEV features are then concatenated with agent-agent interactions (modeled via self-attention in HiVT) followed by a linear projection for the global interaction module to process. As we will show in \cref{sec:expt}, the prediction performance of HiVT and overall system runtime are greatly enhanced by directly encoding BEV features.

\subsection{Augmenting Estimated Lanes with BEV Features}
\label{sec:strategy_2}

While \cref{sec:replace_attend} completely substitutes vectorized map data, another strategy is to 
augment existing lane information with BEV features (e.g., via concatenation). We first refine the BEV features to match the dimensionality of the latent space associated with raw lane information using a one-dimensional CNN. We then determine the BEV grid positions corresponding to the locations of each map vertex and concatenate the original vertex features (i.e., their positions) with their corresponding BEV features. In doing so, we aim to provide a more comprehensive summary of lane information to downstream modules.

In our work, we instantiate this strategy with a combination of the MapTR line of work~\cite{MapTR,maptrv2} and DenseTNT~\cite{GuSunEtAl2021}.  DenseTNT, in particular, is emblematic of prediction models that are heavily map-dependent. It requires lane information at virtually every stage of its pipeline, from initial sparse context encoding to end-point sampling and scoring to guiding predictions during decoding. With such a heavy reliance on vector maps, we cannot eliminate the use of lanes completely as in \cref{sec:replace_attend}. Instead, we focus on enriching the estimated map vertices by incorporating their corresponding BEV features in DenseTNT's input layer. Specifically, the enriched map elements are encoded using a VectorNet~\cite{gao2020vectornet} backbone which we augment with a larger layer size to accommodate the increase in feature dimensionality (full details can be found in the appendix). As we will show in \cref{sec:expt}, incorporating BEV features in this manner significantly improves the performance of the associated behavior predictor. 

\subsection{Replacing Agent Information with Temporal BEV Features}
\label{sec:strategy_3}

Operating from streaming inputs is a common requirement of online mapping methods deployed on embedded devices. Methods like StreamMapNet~\cite{yuan2024streammapnet} introduce a memory buffer to preserve query data and BEV features from previous frames, combining them with the BEV features acquired in the current frame.
This introduction of temporal information into the BEV representation enables our third strategy for incorporating BEV features in behavior prediction: replacing agent information with their corresponding BEV features.

In prediction models, agent trajectories are commonly the only source of temporal information during scene context encoding. Vectorized HD maps provide a static understanding of the scene, with fixed road geometry and semantics. While this static-dynamic separation is explicitly handled by prediction architectures, 
StreamMapNet's approach captures temporal information with a one-step historical BEV feature fusion, enabling it to also capture information about dynamic agents.
To leverage StreamMapNet's encoding of both static \textit{and} dynamic information, we additionally modify DenseTNT~\cite{GuSunEtAl2021} to replace the agent subgraphs encoded in VectorNet~\cite{gao2020vectornet} with the agent-BEV features obtained using the attention mechanism in \cref{eqn:agent_bev_attn}. In doing so, agent trajectory information is completely discarded in DenseTNT~\cite{GuSunEtAl2021}. Even so, as we will show in \cref{sec:expt}, DenseTNT~\cite{GuSunEtAl2021} is able to leverage the implicit trajectory information encoded in the dynamic BEV features and predicts significantly more accurate trajectories.

%% file: sec/4_experiments.tex
\section{Experiments}
\label{sec:expt}

\textbf{Dataset.} We evaluate our method on the large-scale nuScenes dataset~\cite{CaesarBankitiEtAl2019}, which includes ground truth (GT) HD maps, multi-sensor data, and tracked agent trajectories. It consists of 1000 driving scenarios with sensor data recorded at 10 Hz and annotated at 2 Hz (i.e., every 5th frame is a keyframe), and is divided into train, validation, and test sets with 500, 200, and 150 scenarios, respectively.

Our work leverages the unified trajdata~\cite{ivanovic2023trajdata} interface to standardize the data representation between vectorized map estimation models and downstream prediction models. To ensure compatibility across various prediction and mapping models, we upsample the nuScenes trajectory data frequency to 10Hz (matching the sensor frequency) using trajdata's temporal interpolation utilities.
Each prediction model is then tasked with forecasting vehicle motion 3 seconds into the future using observations from the preceding 2 seconds.

\textbf{Metrics.} To evaluate trajectory prediction performance, we adopt standard evaluation metrics used in many recent prediction challenges~\cite{CaesarBankitiEtAl2019,ChangLambertEtAl2019,waymo_open_motion_dataset,wilson2021argoverse2}: minimum Average Displacement Error (minADE), minimum Final Displacement Error (minFDE) and Miss Rate (MR). Specifically, minADE evaluates the average Euclidean ($\ell_2$) distance between the most-accurately predicted trajectory (of 6 predicted trajectories) and the GT trajectory across the prediction horizon. minFDE measures the $\ell_2$ distance between the end point of these two trajectories. MR refers to the proportion of predicted trajectories that have an FDE of more than 2 meters from the GT endpoint.

\textbf{Models and Training.}
To evaluate the effect of incorporating BEV features in downstream prediction models, we train DenseTNT~\cite{GuSunEtAl2021} and HiVT~\cite{zhou2022hivt} on the outputs of four online mapping models (MapTR~\cite{MapTR}, MapTRv2~\cite{maptrv2}, MapTRv2 with Centerlines~\cite{maptrv2}, and StreamMapNet~\cite{yuan2024streammapnet}) in one of three setups: Baseline (using vectorized inputs), Uncertainty-enhanced (where each map element vertex contains spatial uncertainty information)~\cite{GuSongEtAl2024}, and Ours (one of the BEV feature attention strategies detailed in \cref{sec:methods}), yielding a total of 24 model combinations.

In particular, we first train the four online map estimation models to convergence following the models' original training recipes in the baseline setup or the training recipes of~\cite{GuSongEtAl2024} in the uncertainty-enhanced setup. We then extract BEV features from each model and scene, and train behavior prediction models according to the above three setups: using only vectorized map information as a baseline, leveraging uncertainty as in~\cite{GuSongEtAl2024}, and our approach of leveraging BEV features as in~\cref{sec:methods}. All models are trained on a single RTX4090 GPU. Full model and training details can be found in the appendix.

\subsection{Leveraging BEV Features in Behavior Prediction}

\input{sec/quant_table}

\textbf{Prediction Accuracy Improvements.}
As shown in \cref{tab:quant}, for virtually all mapping/prediction model combinations, incorporating BEV features leads to \textit{significantly} better prediction accuracy, not only compared to the baseline models but also to the uncertainty-enhanced approach. The largest improvements (up to $25\%$ and more) are in MR and minFDE, suggesting that latent BEV features can especially help with long-horizon prediction performance. Endpoint prediction accuracy is especially important for trajectory prediction as it directly impacts later planning accuracy.

Note that, while MapTR~\cite{MapTR} does not perform as well as its successor (MapTRv2~\cite{maptrv2}) or the temporally-enhanced StreamMapNet~\cite{yuan2024streammapnet}, its BEV features provide the largest improvement to downstream performance, yielding prediction accuracy that outperforms combinations with MapTRv2~\cite{maptrv2} (without centerlines). This result suggests that MapTR's decoder may be introducing unwanted noise, which impedes its ability to generate precise map elements. Accordingly, by leveraging information from earlier stages of MapTR~\cite{MapTR}, both HiVT~\cite{zhou2022hivt} and DenseTNT~\cite{GuSunEtAl2021} obtain significant improvements in prediction performance, emphasizing the benefits of deeper integrations between mapping and prediction through an intermediate BEV representation.

The additional production of centerlines in MapTRv2-CL~\cite{maptrv2} yields the most accurate predictions overall. Accordingly, the benefits of incorporating BEV features are the least pronounced. This result reaffirms the utility of centerlines for trajectory prediction and provides guidance for future map estimation research regarding which type of map element is most useful. 

Finally, although StreamMapNet's BEV features are leveraged in two completely different ways (as a substitute for lane information in HiVT~\cite{zhou2022hivt} and as a substitute for agent information in DenseTNT~\cite{GuSunEtAl2021}), they both provide significant improvements to the baseline prediction models. This indicates that temporal information not only helps in decoding more accurate maps, it also provides a temporal understanding of the behavior of agents which is particularly valuable for trajectory prediction.

\begin{figure}[t]
    \centering
    \includegraphics[width=0.45\linewidth]{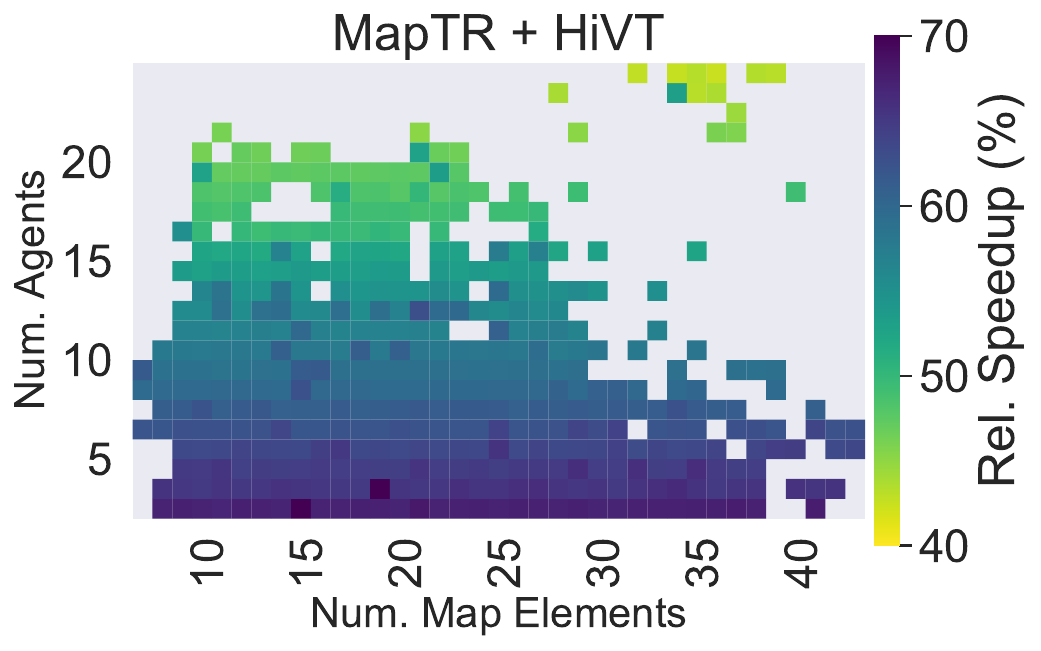}
    \hfill
    \includegraphics[width=0.45\linewidth]{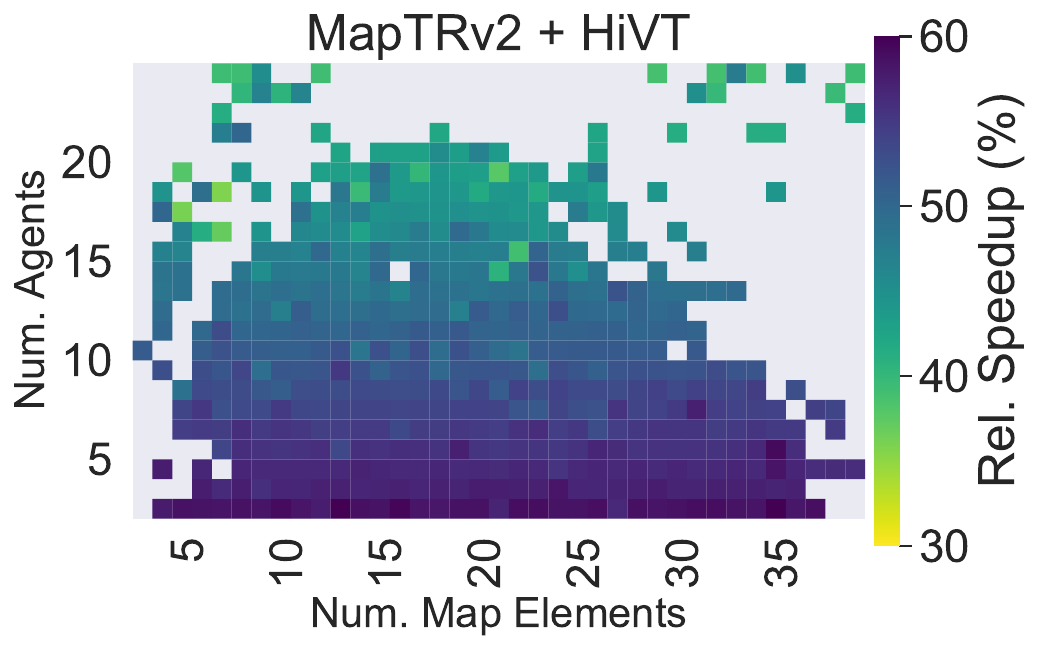}
    \includegraphics[width=0.8\linewidth]{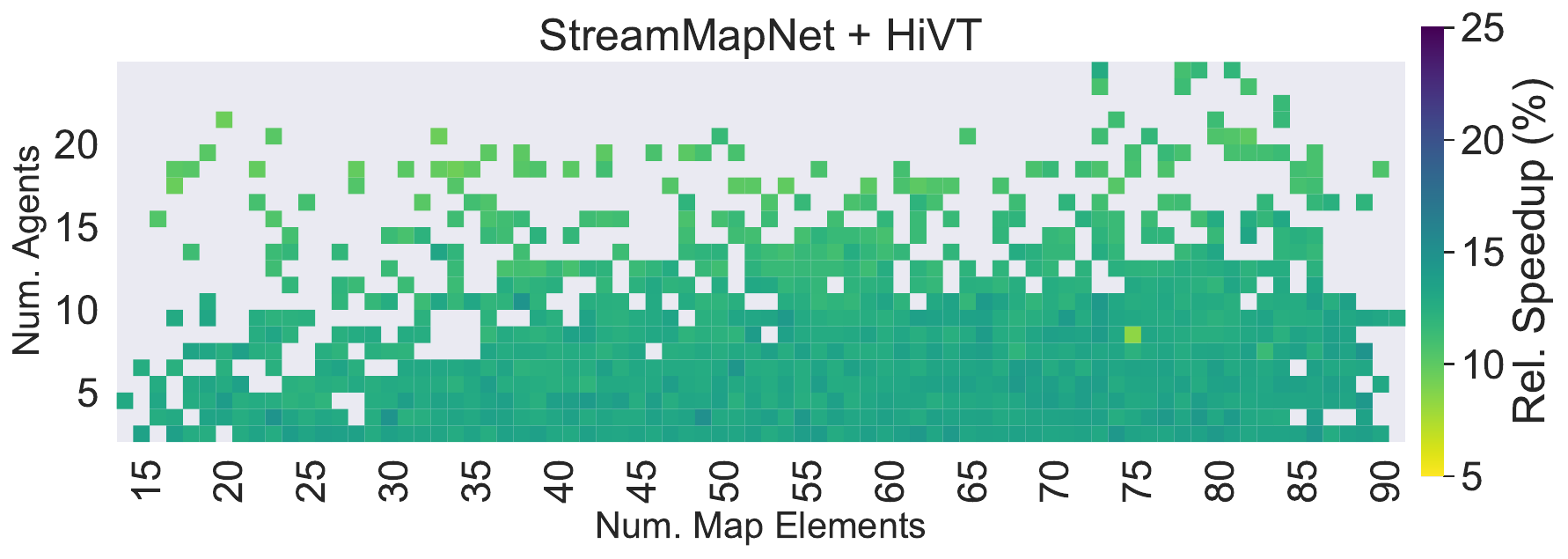}

    \vspace{-0.3cm}
    
    \caption{Our integrated BEV-prediction approach runs faster than decoupled baselines across all scenario sizes (number of agents and map elements) and mapping models.}
    \label{fig:runtime}

    \vspace{-0.5cm}
\end{figure}

\textbf{Inference Speedup.} In \cref{fig:runtime}, we compare the GPU inference speedup achieved by our integrated approach (described in \cref{sec:replace_attend}) relative to decoupled baselines for HiVT~\cite{zhou2022hivt}. For both approaches, runtime is measured starting from the processing of the input RGB images and ending at the output of the final trajectories.
As can be seen, our approach results in significant inference time improvements due to its elimination of the time-consuming map decoding stage. Specifically, our integrated method is 42-73\% faster than MapTR~\cite{MapTR} and HiVT\cite{zhou2022hivt} alone, 35-62\% faster than MapTRv2~\cite{maptrv2} and HiVT\cite{zhou2022hivt} alone, and 8-15\% faster than StreamMapNet~\cite{yuan2024streammapnet} and HiVT\cite{zhou2022hivt} alone. Identical improvements are obtained when compared to the uncertainty-integrated method~\cite{GuSongEtAl2024}, reaffirming the strength of our approach.

Further, the overall inference time of the baseline mapping and prediction models scales with both the number of map elements and number of agents present in a scene. In contrast, our integrated approach is much less sensitive to the number of map elements, and this is reflected in \cref{fig:runtime} where our approach's runtime improves more as the number of map elements increases.
This occurs because each agent's patch attends to every other BEV \textit{patch} in \cref{sec:replace_attend}. Thus, even if the number of map elements is high, the number of BEV patches is fixed, significantly reducing inference time by eliminating the need to process map elements through HiVT's encoder.

In scenarios where the number of map elements is low but the number of agents is high (top left of \cref{fig:runtime}), the reduction in processing time is less pronounced. Given the smaller quantity of map elements, HiVT~\cite{zhou2022hivt} naturally requires less time to encode them. Further, with an increase in the number of agents, the approach in \cref{sec:replace_attend} must perform additional attention operations as each new agent introduces an extra $N$ attention operations. This increase in computation partially offsets the savings achieved by not processing vectorized map elements, leading to smaller improvements in run time.
Nevertheless, our approach still yields substantial reductions in inference time. 

\subsection{Ablation Studies}
\label{sec:abalation}

\textbf{Patch Size.} \cref{tab:patch} illustrates the effect of BEV feature patch size $P$ on prediction performance. A patch size that is too small results in insufficient information capture. For instance, MapTRv2 operates with a $60m \times 30m$ perception range and has a BEV dimension ($H \times W$) of $200 \times 100$, meaning each BEV grid cell represents a $0.3m \times 0.3m$ square in the real world. This size is relatively small, particularly as we wish to capture global information in the scene. As shown in \cref{tab:patch}, a patch size of $10 \times 5$ (covering $3m \times 1.5m$) underperforms compared to a more moderately-sized $20 \times 20$ patch ($6m \times 6m$). Larger patch sizes do not yield performance improvements, however, likely due to the loss of granular information when converting patches into smaller vector embeddings (projecting from $P^2 D$ to $D$ dimensions), which also inhibits prediction accuracy as reflected by worsening minFDE and MR values. Overall, we find that a patch size of $20 \times 20$ yields the best prediction performance.

\begin{table}[tb]
  \centering
  \setlength{\tabcolsep}{4pt} %
  \renewcommand{\arraystretch}{0.8} %
  \begin{tabular}{@{}c|ccc@{}}
    \toprule
    Models & \multicolumn{3}{c}{MapTRv2-Centerline~\cite{maptrv2} and HiVT~\cite{zhou2022hivt}} \\
    \midrule
    Patch Sizes & minADE $\downarrow$ & minFDE $\downarrow$ & MR $\downarrow$ \\
    \midrule
    (10, 5) & 0.3845 & 0.8003 & 0.0853 \\
    (20, 10) & 0.3729 & 0.7616 & 0.0749 \\
    (20, 20) & 0.3728 & \textbf{0.7518} & \textbf{0.0737} \\
    (20, 25) & \textbf{0.3709} & 0.7649 & 0.0761 \\
    (40, 20) & 0.3737 & 0.7583 & 0.0770 \\
    \bottomrule
  \end{tabular}
  \vspace{0.1cm}
  \caption{An exploration of BEV patch sizes reveals that there are detriments to having patches that are too small (insufficient information capture) or too large (granular information loss), with the best performance achieved by a patch size of $20 \times 20$ (corresponding to $6m \times 6m$ in the real world).}
  \label{tab:patch}
  
  \vspace{-0.8cm} %
\end{table}

\textbf{BEV Encoder Selection.} As mentioned in ~\cref{sec:related_work}, map estimation models typically employ one of two distinct PV2BEV encoders: BEVFormer~\cite{li2022bevformer}, which is used in MapTR~\cite{MapTR} and StreamMapNet~\cite{yuan2024streammapnet}, and LSS~\cite{philion2020lss}, used in MapTRv2~\cite{maptrv2} and MapTRv2-Centerline~\cite{maptrv2}. As described in \cref{sec:methods}, BEVFormer~\cite{li2022bevformer} enhances its BEV features by integrating historical BEV features \((B_{t-1})\) through temporal self-attention. In contrast, LSS~\cite{philion2020lss} processes only data from the current frame. This divergence in backbone mechanisms leads to notable performance disparities between MapTR~\cite{MapTR} and its subsequent derivatives MapTRv2~\cite{maptrv2} and MapTRv2-Centerline~\cite{maptrv2}. Despite the use of a similar decoding mechanism (a hierarchical query embedding scheme followed by hierarchical bipartite matching),
the utilization of an LSS-based BEV extractor results in a lack of temporal information in MapTRv2's resulting BEV features. This results in relative inefficiency within our methodology, showcasing only modest enhancements when compared against integrations with MapTR~\cite{MapTR} and StreamMapNet~\cite{yuan2024streammapnet}. 

This divergence is empirically observed in \cref{tab:quant}, where MapTRv2 and MapTRv2-Centerline's BEV features (which only encode static information) yield a mere relative improvement of $4\%$, $2\%$ and $4\%$ in minADE, minFDE, and MR, respectively. In comparison, the use of temporally-informed BEV features in MapTR~\cite{MapTR} and StreamMapNet~\cite{yuan2024streammapnet} yields relative performance enhancements of $16\%$, $19\%$, and $24\%$ in minADE, minFDE, and MR for both mapping/prediction model combinations. These improvements underscore the utility of integrating temporal dynamics into BEV features for improved trajectory forecasting. 

\input{sec/qua_figs}

\subsection{Qualitative Comparisons}
\label{sec:qual}

\textbf{BEV Feature Visualization.} Aside from estimated maps and their uncertainties,
we also visualize the corresponding BEV features for each test scene in \cref{fig:vis1,fig:vis2,fig:vis3}. These visualizations are obtained by first reducing the dimensionality of each BEV grid cell to a single value using principal component analysis (PCA), followed by normalization to $[0, 255]$, creating a grayscale image. The resulting BEV feature images form the background of our method plots.

In instances where SteamMapNet~\cite{yuan2024streammapnet} serves as the mapping model (\cref{fig:vis1,fig:vis3}), a distinct separation is observed between driveable areas (gray) and the non-driveable areas (white) beyond the designated boundaries. This distinction highlights the comprehensive geometric information captured within the temporal BEV features of SteamMapNet~\cite{yuan2024streammapnet}, enabling the implicit modeling of map characteristics from BEV features alone. This interpretation is inverted in MapTR's BEV features (\cref{fig:vis2}), white indicates driveable areas and gray signifies non-driveable regions. In both cases, BEV features play a critical role in informing future predictions.

\cref{fig:vis1} visualizes a parking lot next to a building. Utilizing the temporal BEV features from StreamMapNet~\cite{yuan2024streammapnet}, HiVT~\cite{zhou2022hivt} is able to have all six predicted trajectories tightly clustered around the GT. This precision allows HiVT~\cite{zhou2022hivt} to maintain lane discipline, avoiding encroaching into opposing lanes (seen in the Baseline and Uncertainty setups).

In \cref{fig:vis2}, MapTR's estimated map vertices are augmented with their corresponding BEV features. Specifically, the grey area depicted ahead of the center vehicle covers part of the road boundary, which provides an additional non-driveable signal and reinforces the presence of a road border. By leveraging this extra information, DenseTNT~\cite{GuSunEtAl2021} much more effectively restricts the vehicle's trajectories from crossing into the non-drivable area. In contrast, both the Baseline and Uncertainty-enhanced approaches produce trajectories that intersect with the road boundary and lane divider. By incorporating BEV features, DenseTNT~\cite{GuSunEtAl2021} is able to confine its future predictions to the designated white, driveable area, showing the utility of BEV features in improving map adherence.

In \cref{fig:vis3}, the Baseline predicted trajectories undershoot the GT, whereas the Uncertainty-enhanced approach overshoots the pedestrian crosswalk entirely. 
By encoding the BEV features directly for agent information, our approach strikes a balance and produces trajectories that align well with the GT trajectory.

%% file: sec/quant_table.tex
\begin{table*}[tb]
  \centering
  \setlength{\tabcolsep}{3pt}
  \resizebox{1\linewidth}{!}{
  \begin{tabular}{@{}l|lll|lll@{}}
    \toprule
    Prediction Method & \multicolumn{3}{c|}{HiVT~\cite{zhou2022hivt}} & \multicolumn{3}{c}{DenseTNT~\cite{GuSunEtAl2021}} \\
    
    \midrule
    Online HD Map Method & minADE $\downarrow$ & minFDE $\downarrow$ & MR $\downarrow$ & minADE $\downarrow$ & minFDE $\downarrow$ & MR $\downarrow$ \\
    
    \midrule
    MapTR~\cite{MapTR} & 0.4234 & 0.8900 & 0.0955 & 1.0462 & 2.0661 & 0.3494 \\
    MapTR~\cite{MapTR} + Unc~\cite{GuSongEtAl2024} & 0.4036 & 0.8372 & 0.0822 & 1.1190 & 2.1502 & 0.3669 \\
    MapTR~\cite{MapTR} + Ours & \textbf{0.3617} {\small \color{darkpastelgreen} ($\mathbf{-15}\%$)} & \textbf{0.7401} {\small \color{darkpastelgreen} ($\mathbf{-17}\%$)} & \textbf{0.0720} {\small \color{darkpastelgreen} ($\mathbf{-25}\%$)} & \textbf{0.7608} {\small \color{darkpastelgreen} ($\mathbf{-27}\%$)} & \textbf{1.4700} {\small \color{darkpastelgreen} ($\mathbf{-29}\%$)}  & \textbf{0.2593} {\small \color{darkpastelgreen} ($\mathbf{-26}\%$)} \\
    
    \midrule
    MapTRv2~\cite{maptrv2} & 0.3950 & 0.8310 & 0.0894 & 1.2648 & 2.3481 & 0.4043 \\ 
    MapTRv2~\cite{maptrv2} + Unc~\cite{GuSongEtAl2024} & 0.3896  & 0.8085 & 0.0859  & 1.3228 & 2.4821 & 0.4406  \\
    MapTRv2~\cite{maptrv2} + Ours & \textbf{0.3844} {\small \color{darkpastelgreen} ($-3\%$)} & \textbf{0.7848} {\small \color{darkpastelgreen} ($-6\%$)} & \textbf{0.0741} {\small \color{darkpastelgreen} ($\mathbf{-17}\%$)} & \textbf{1.1232} {\small \color{darkpastelgreen} ($\mathbf{-11}\%$)} & \textbf{2.3000} {\small \color{darkpastelgreen} ($-2\%$)} & \textbf{0.4025} {\small \color{gray} ($0\%$)} \\
    
    \midrule
    MapTRv2-CL~\cite{maptrv2} & 0.3657 & 0.7473 & 0.0710 & 0.7664 & \textbf{1.3174} & \textbf{0.1547}  \\
    MapTRv2-CL~\cite{maptrv2} + Unc~\cite{GuSongEtAl2024} & \textbf{0.3588} & \textbf{0.7232} & \textbf{0.0660} & 0.8123 & 1.3426 & 0.1567  \\
    MapTRv2-CL~\cite{maptrv2} + Ours & 0.3652 {\small \color{gray} ($0\%$)} & 0.7323 {\small \color{darkpastelgreen} ($-2\%$)} & 0.0710 {\small \color{gray} ($0\%$)} & \textbf{0.7630} {\small \color{gray} ($0\%$)} & 1.3609 {\small \color{orange} ($+3\%$)} & 0.1576 {\small \color{orange} ($+2\%$)} \\
    
    \midrule
    StreamMapNet~\cite{yuan2024streammapnet}  & 0.4035 & 0.8569 & 0.0996 & 0.8864 & 1.7050 & 0.2467 \\
    StreamMapNet~\cite{yuan2024streammapnet}  + Unc~\cite{GuSongEtAl2024} & 0.3907 & 0.8034 & 0.0812 & 0.9220 & 1.6851 & 0.2310 \\
    StreamMapNet~\cite{yuan2024streammapnet}  + Ours & \textbf{0.3800} {\small \color{darkpastelgreen} ($-6\%$)} & \textbf{0.7709} {\small \color{darkpastelgreen} ($\mathbf{-10}\%$)} & \textbf{0.0746} {\small \color{darkpastelgreen} ($\mathbf{-25}\%$)} & \textbf{0.7377} {\small \color{darkpastelgreen} ($\mathbf{-17}\%$)} & \textbf{1.3661} {\small \color{darkpastelgreen} ($\mathbf{-20}\%$)} & \textbf{0.1987} {\small \color{darkpastelgreen} ($\mathbf{-19}\%$)} \\
    
    \bottomrule
  \end{tabular}
  }

  \vspace{0.1cm}
  
  \caption{
  Virtually every combination of mapping and prediction benefits from directly leveraging upstream BEV features on the 
  nuScenes~\cite{CaesarBankitiEtAl2019} dataset, with certain combinations achieving performance improvements of 25\% or more.
  Percent values denote the relative improvement in prediction performance achieved by our approach.}
  \label{tab:quant}

  \vspace{-0.8cm}

\end{table*}

%% file: sec/qua_figs.tex
\begin{figure*}[t]
\centering

\begin{minipage}[b]{.24\textwidth}
    \includegraphics[width=\linewidth]{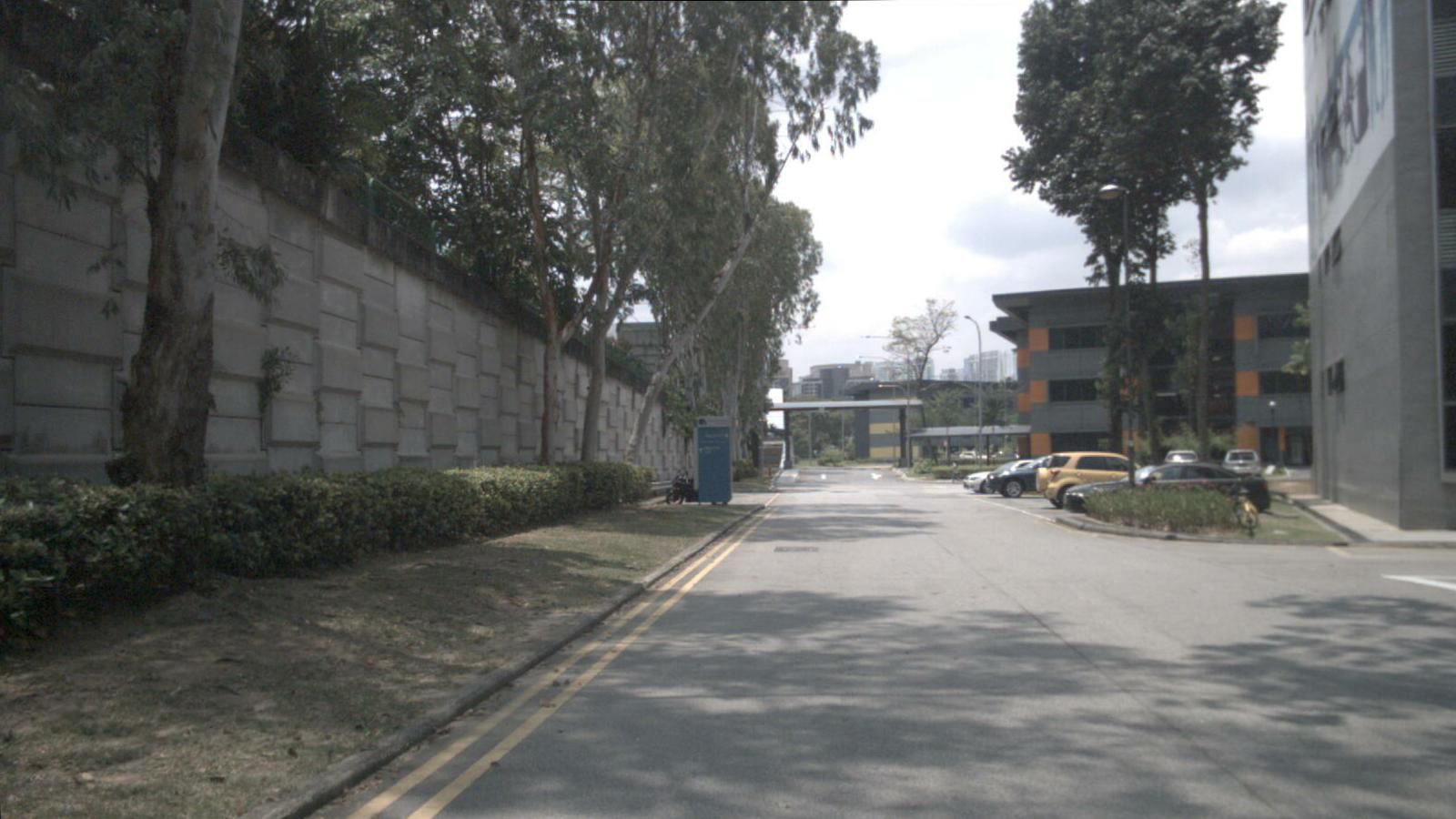}
    \includegraphics[width=\linewidth]{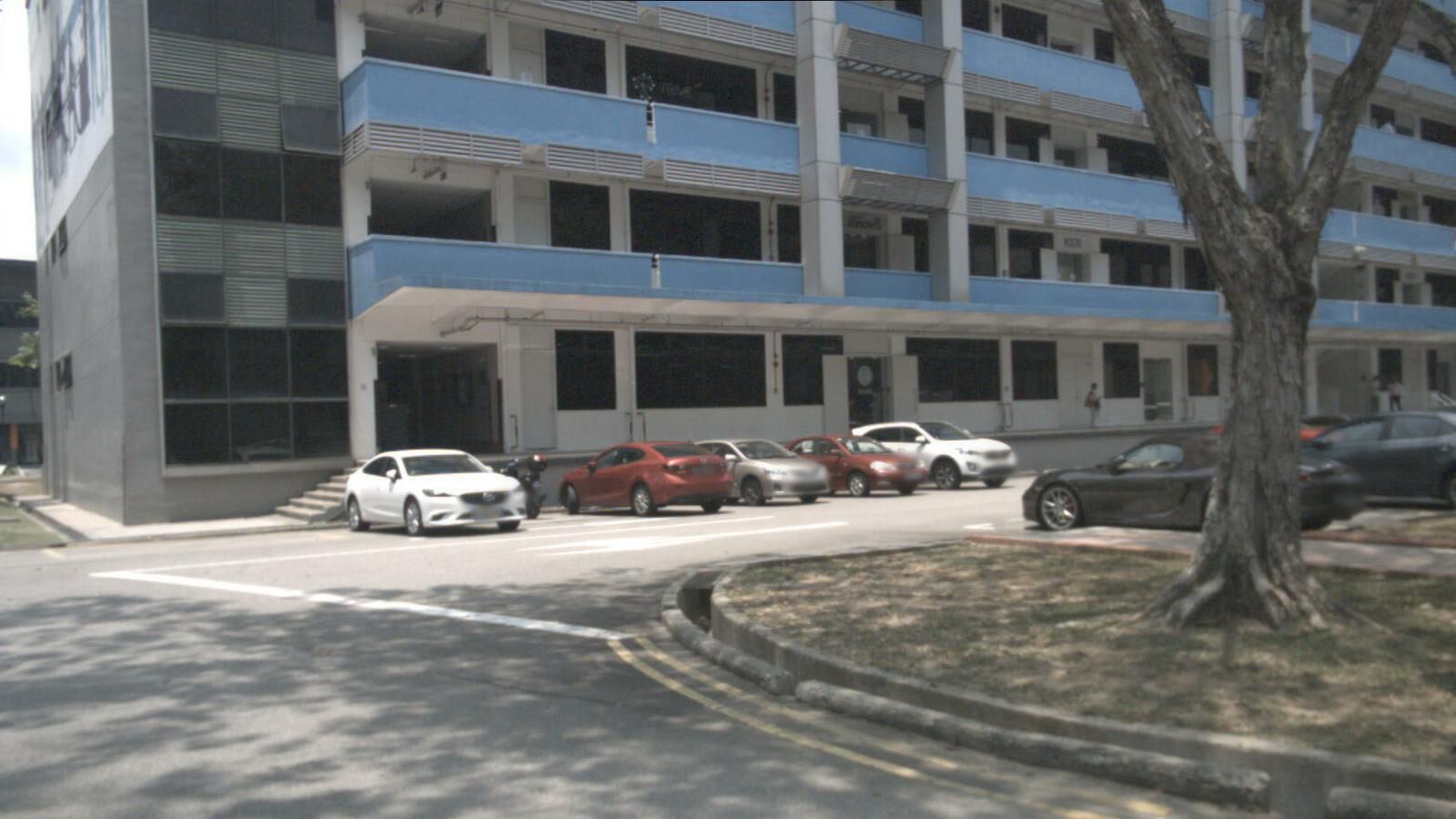}
    \includegraphics[width=\linewidth]{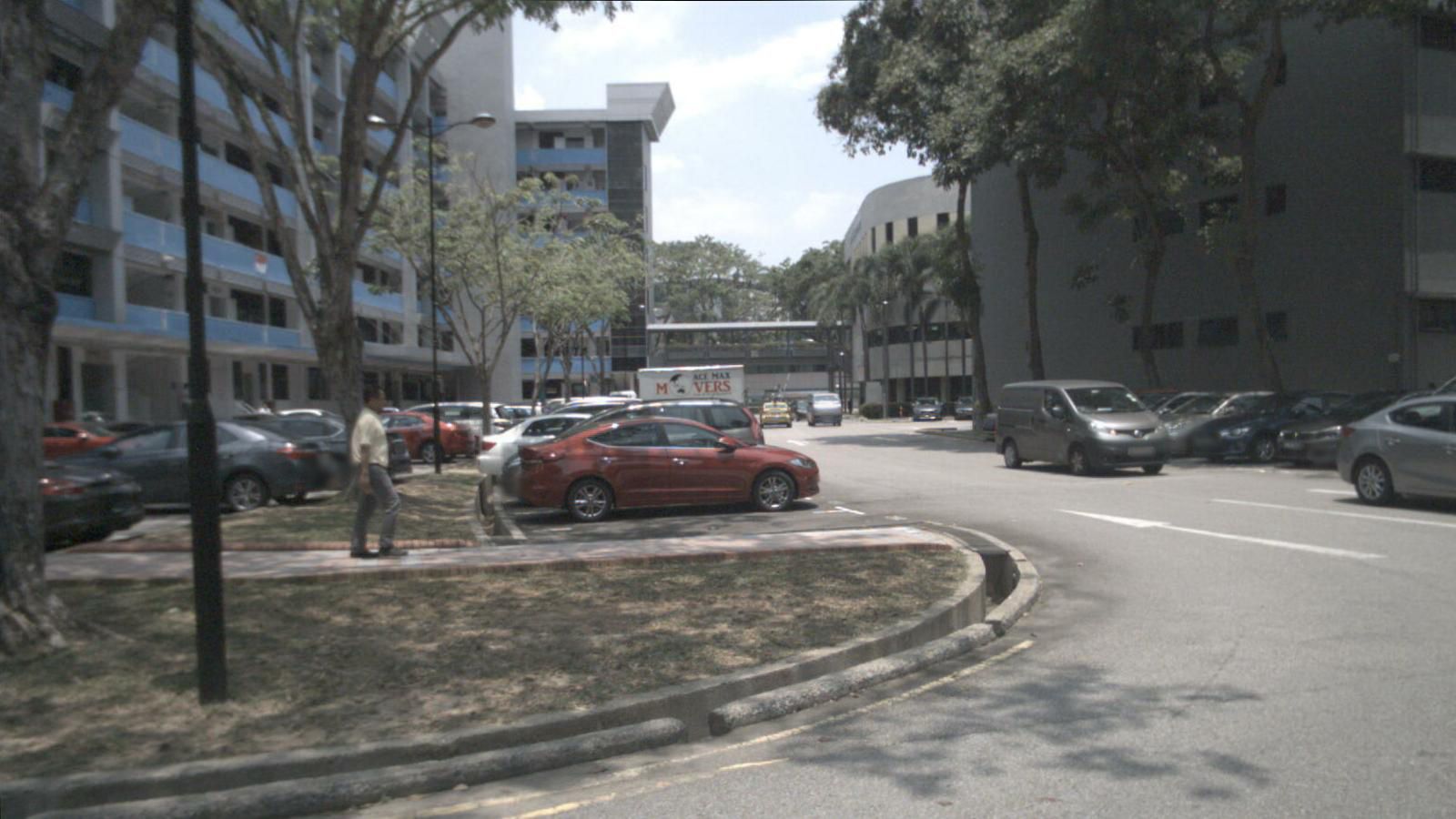}
\end{minipage}%
\hfill %
\begin{minipage}[b]{.74\textwidth}
  \centering
  \begin{subfigure}{.24\linewidth}
    \includegraphics[width=\linewidth]{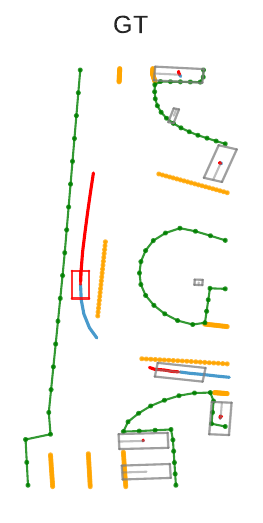}
    \label{fig:vis1_gt}
  \end{subfigure}%
  \begin{subfigure}{.24\linewidth}
    \includegraphics[width=\linewidth]{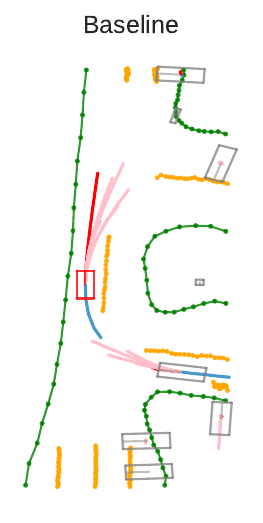}
    \label{fig:vis1_hmn}
  \end{subfigure}%
  \begin{subfigure}{.24\linewidth}
    \includegraphics[width=\linewidth]{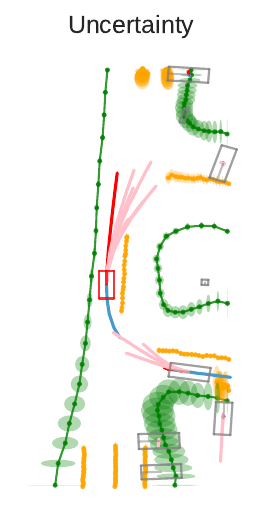}
    \label{fig:vis1_hmu}
  \end{subfigure}%
  \begin{subfigure}{.24\linewidth}
    \includegraphics[width=\linewidth]{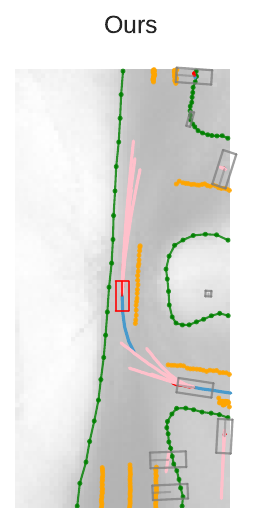}
    \label{fig:vis1_dsu}
  \end{subfigure}

    \vspace{-0.4cm}

    \begin{subfigure}{0.92\linewidth}
        \begin{tikzpicture}
        \draw [darkgreen, ultra thick] (0,0.5) -- (0.35,0.5) node[right, black] {\scriptsize Road Boundary};
        \draw [darkyellow, ultra thick] (2.65,0.5) -- (3,0.5) node[right, black] {\scriptsize Lane Divider};
        \draw [blue, ultra thick] (5.0,0.5) -- (5.35,0.5) node[right, black] {\scriptsize Pedestrian Crossing};
        \draw [red, ultra thick] (0,0) -- (0.35,0) node[right, black] {\scriptsize GT Future};
        \draw [pink, ultra thick] (2.1,0) -- (2.45,0) node[right, black] {\scriptsize Predicted Trajectories};
        \draw [historyblue, ultra thick] (5.65,0) -- (6,0) node[right, black] {\scriptsize Agent History};
        \end{tikzpicture}
  \end{subfigure}

\end{minipage}

\vspace{-0.2cm}

\caption{StreamMapNet~\cite{yuan2024streammapnet} and HiVT~\cite{zhou2022hivt} combined using the strategy in \cref{sec:replace_attend}. By replacing lane information with temporal BEV features, HiVT is able to keep its predicted trajectories in the current lane, closely aligning with the GT trajectory.}
\label{fig:vis1}

\vspace{-0.31cm}

\end{figure*}
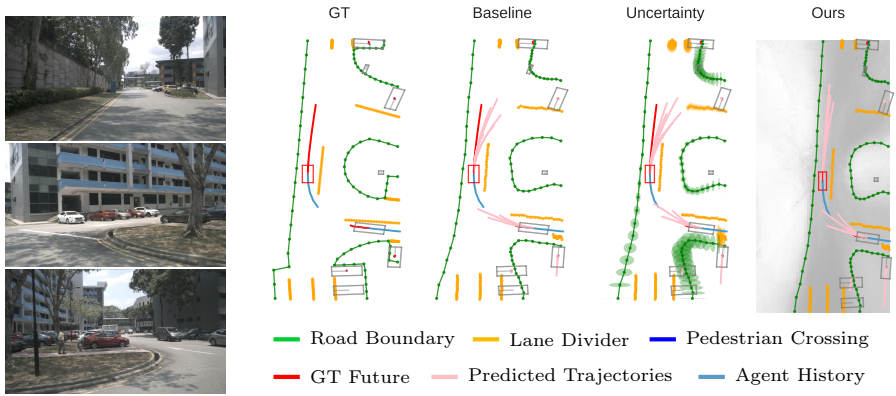

\begin{figure*}[t]
\centering

\begin{minipage}[b]{.24\textwidth}
    \includegraphics[width=\linewidth]{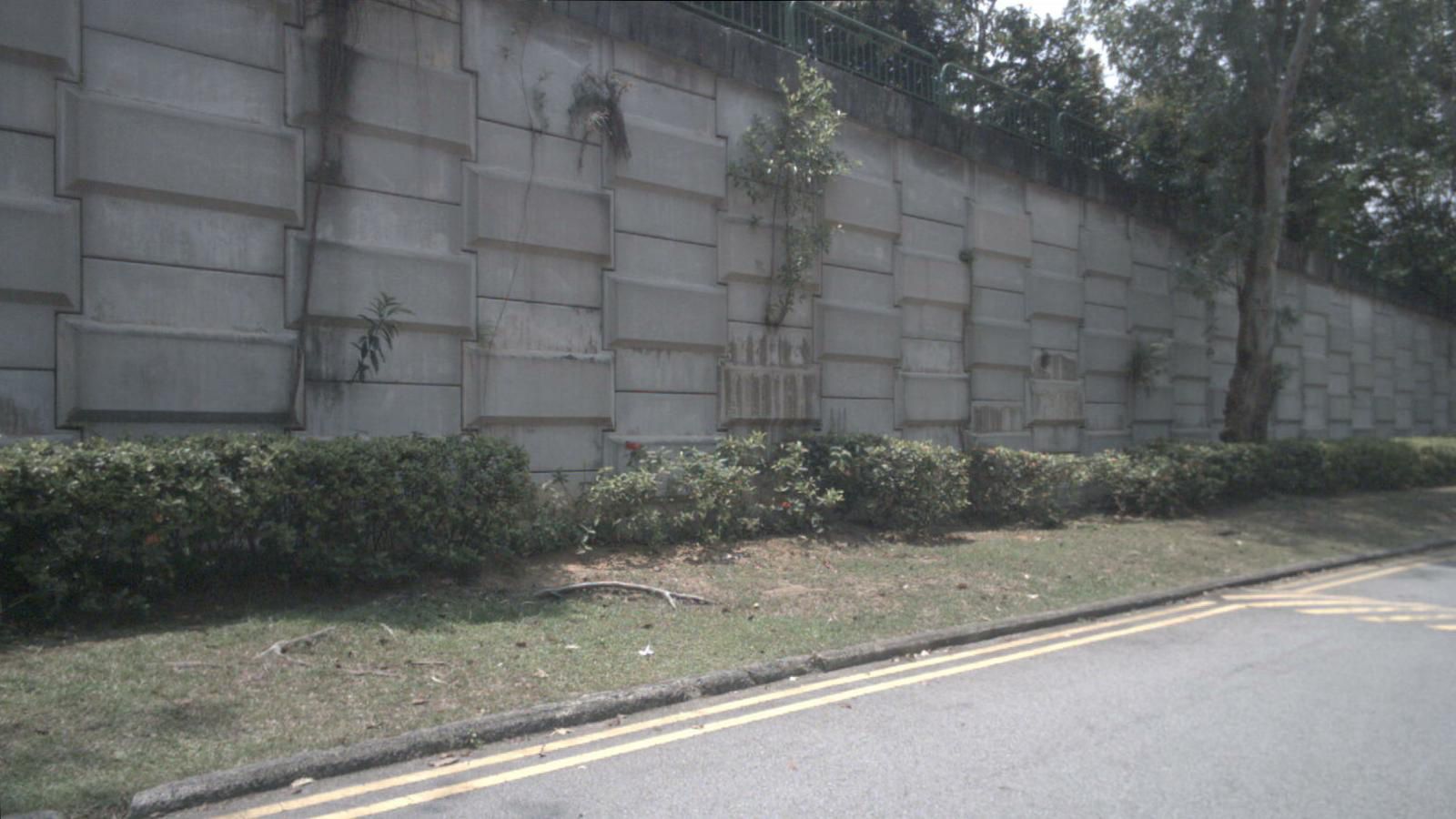}
    \includegraphics[width=\linewidth]{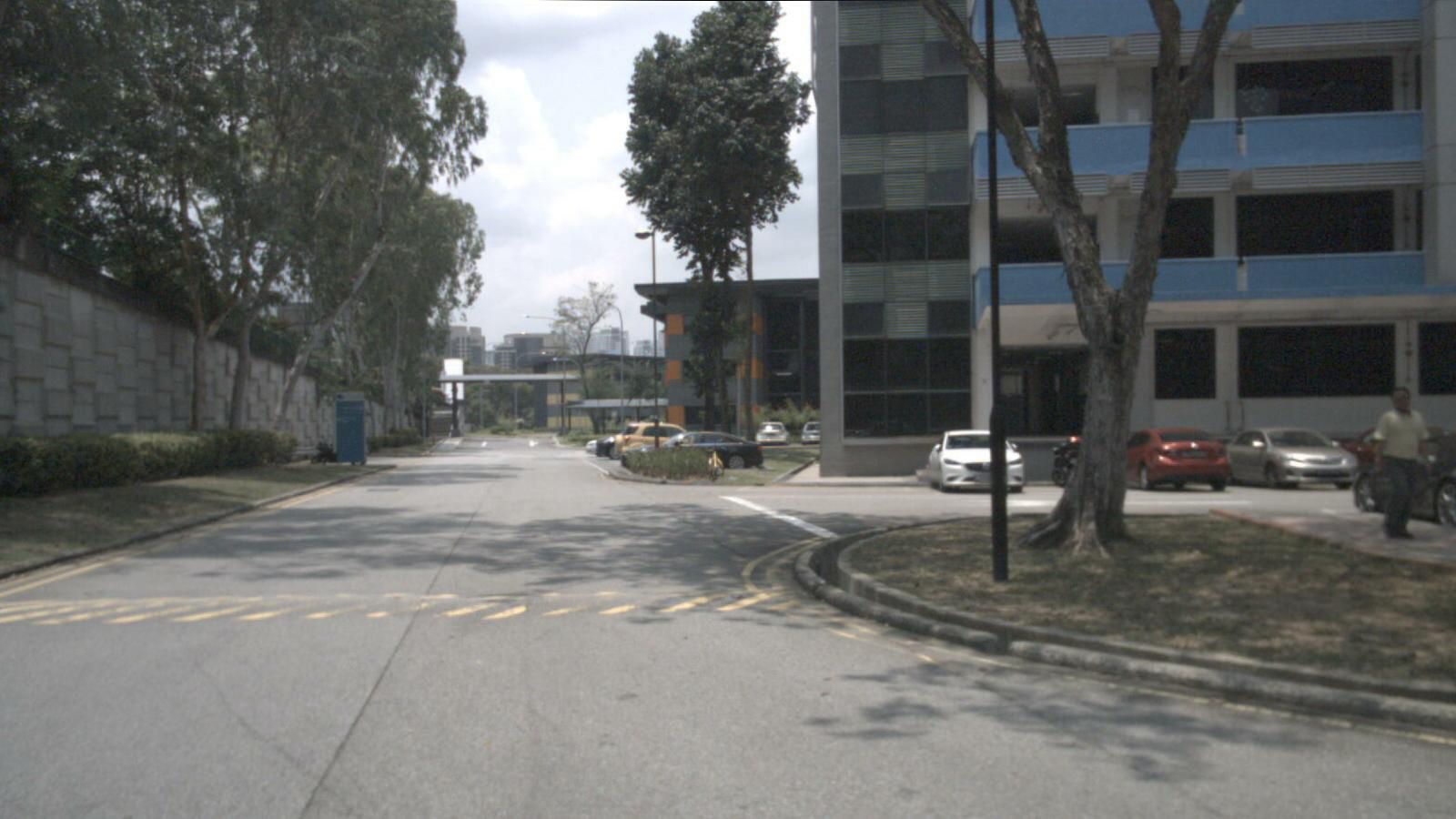}
    \includegraphics[width=\linewidth]{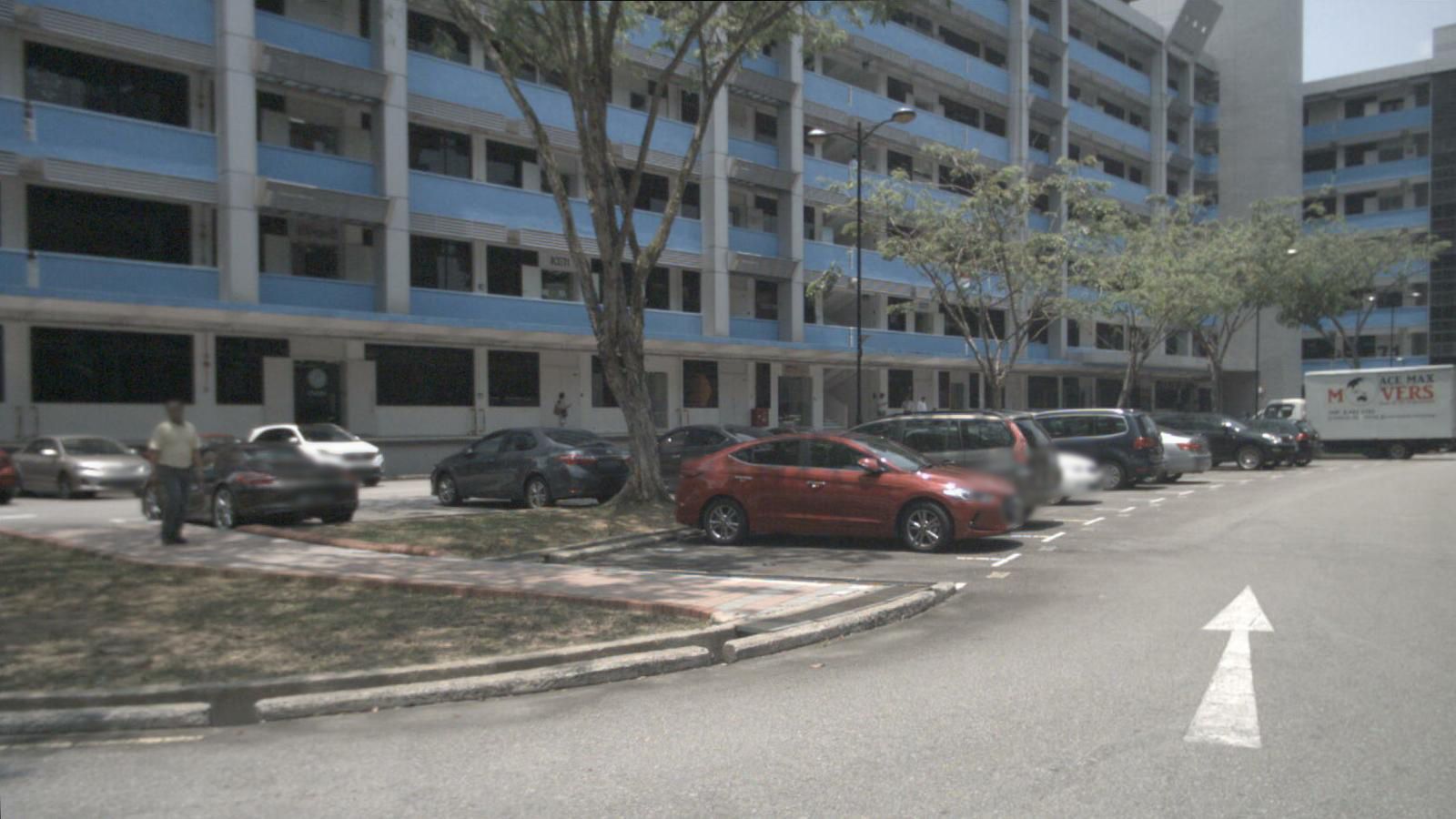}
\end{minipage}%
\hfill %
\begin{minipage}[b]{.74\textwidth}
  \centering
  \begin{subfigure}{.24\linewidth}
    \includegraphics[width=\linewidth]{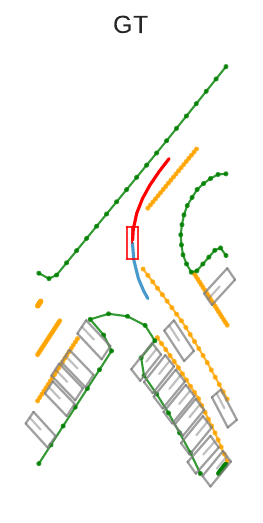}
    \label{fig:vis2_gt}
  \end{subfigure}%
  \begin{subfigure}{.24\linewidth}
    \includegraphics[width=\linewidth]{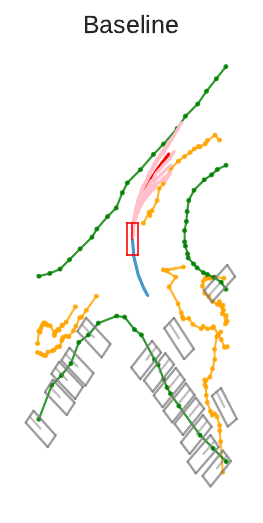}
    \label{fig:vis2_hmn}
  \end{subfigure}%
  \begin{subfigure}{.24\linewidth}
    \includegraphics[width=\linewidth]{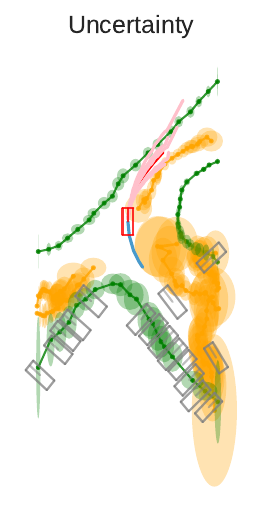}
    \label{fig:vis2_hmu}
  \end{subfigure}%
  \begin{subfigure}{.24\linewidth}
    \includegraphics[width=\linewidth]{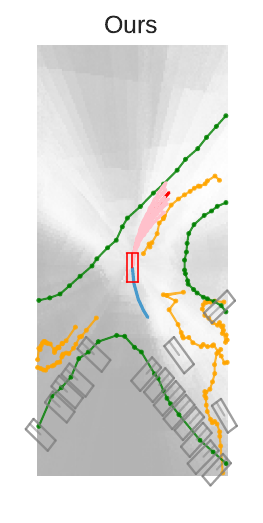}
    \label{fig:vis2_dsu}
  \end{subfigure}

    \vspace{-0.4cm}

\begin{subfigure}{0.92\linewidth}
    \begin{tikzpicture}
    \draw [darkgreen, ultra thick] (0,0.5) -- (0.35,0.5) node[right, black] {\scriptsize Road Boundary};
    \draw [darkyellow, ultra thick] (2.65,0.5) -- (3,0.5) node[right, black] {\scriptsize Lane Divider};
    \draw [blue, ultra thick] (5.0,0.5) -- (5.35,0.5) node[right, black] {\scriptsize Pedestrian Crossing};
    \draw [red, ultra thick] (0,0) -- (0.35,0) node[right, black] {\scriptsize GT Future};
    \draw [pink, ultra thick] (2.1,0) -- (2.45,0) node[right, black] {\scriptsize Predicted Trajectories};
    \draw [historyblue, ultra thick] (5.65,0) -- (6,0) node[right, black] {\scriptsize Agent History};
    \end{tikzpicture}
\end{subfigure}
\end{minipage}

\vspace{-0.2cm}

\caption{MapTR~\cite{MapTR} and DenseTNT~\cite{GuSunEtAl2021} combined via the strategy in \cref{sec:strategy_2}. Our augmentation of map vertices with BEV features enables DenseTNT to produce very accurate trajectories, preventing the road boundary incursions seen in the Baseline and Uncertainty-enhanced~\cite{GuSongEtAl2024} setups.}
\label{fig:vis2}

\vspace{-0.5cm}

\end{figure*}
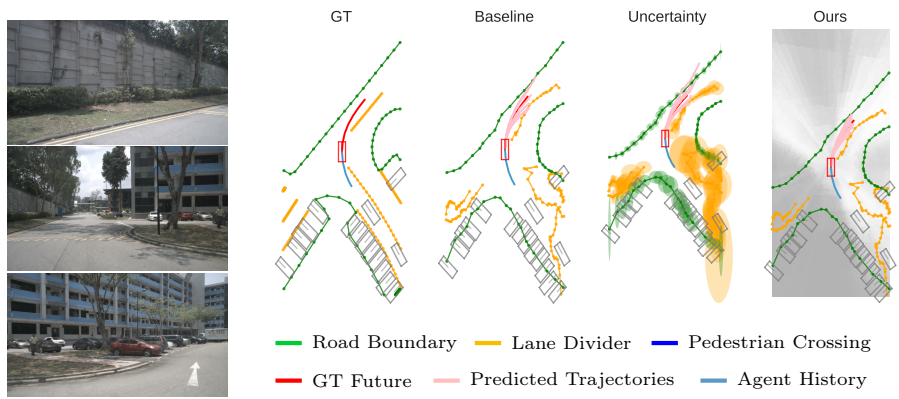

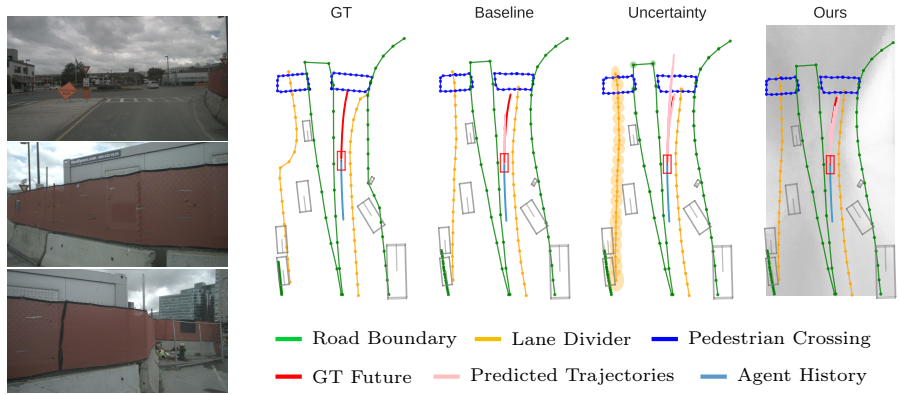
\begin{figure*}[t]
\centering

\begin{minipage}[b]{.24\textwidth}
    \includegraphics[width=\linewidth]{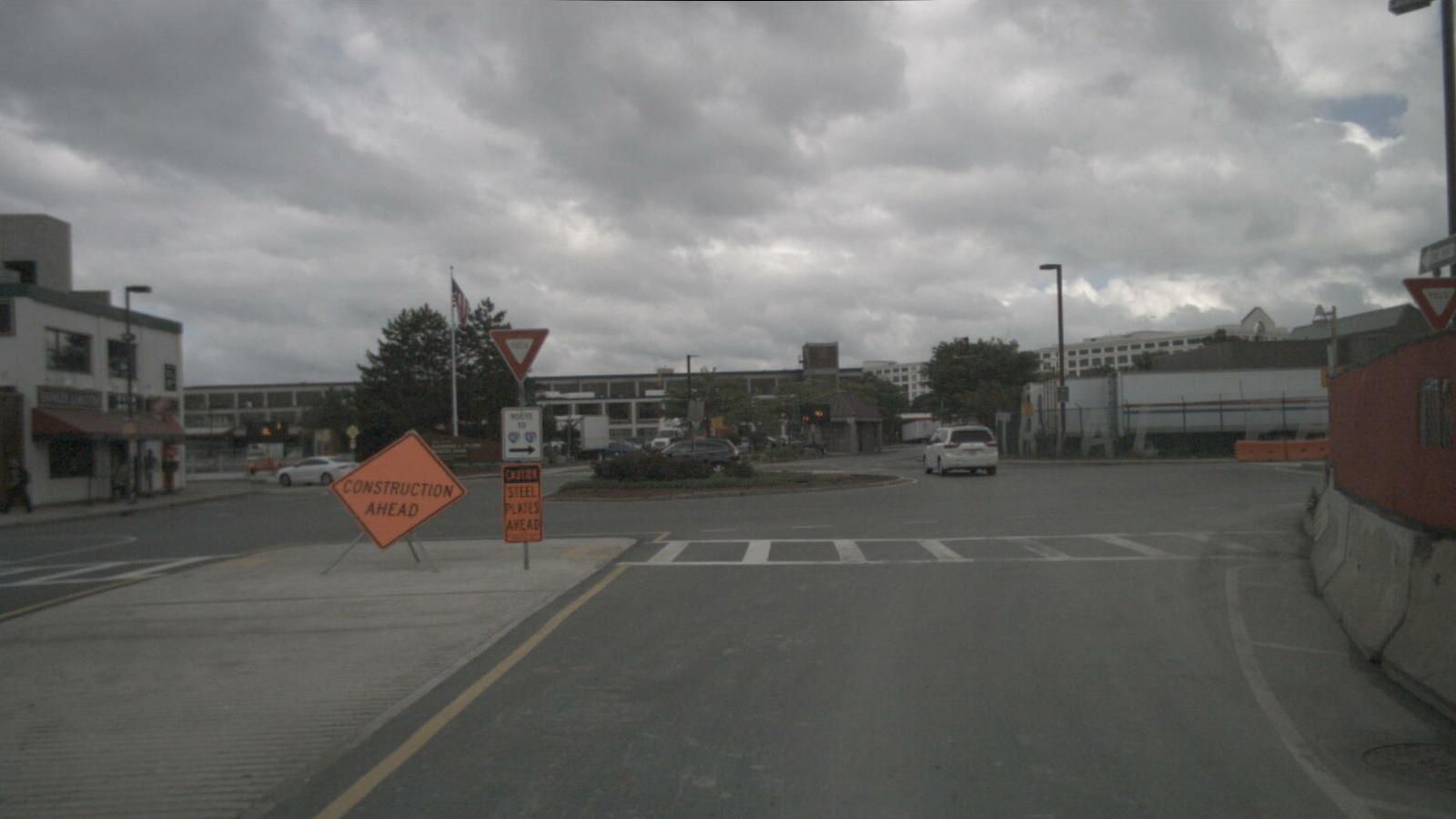}
    \includegraphics[width=\linewidth]{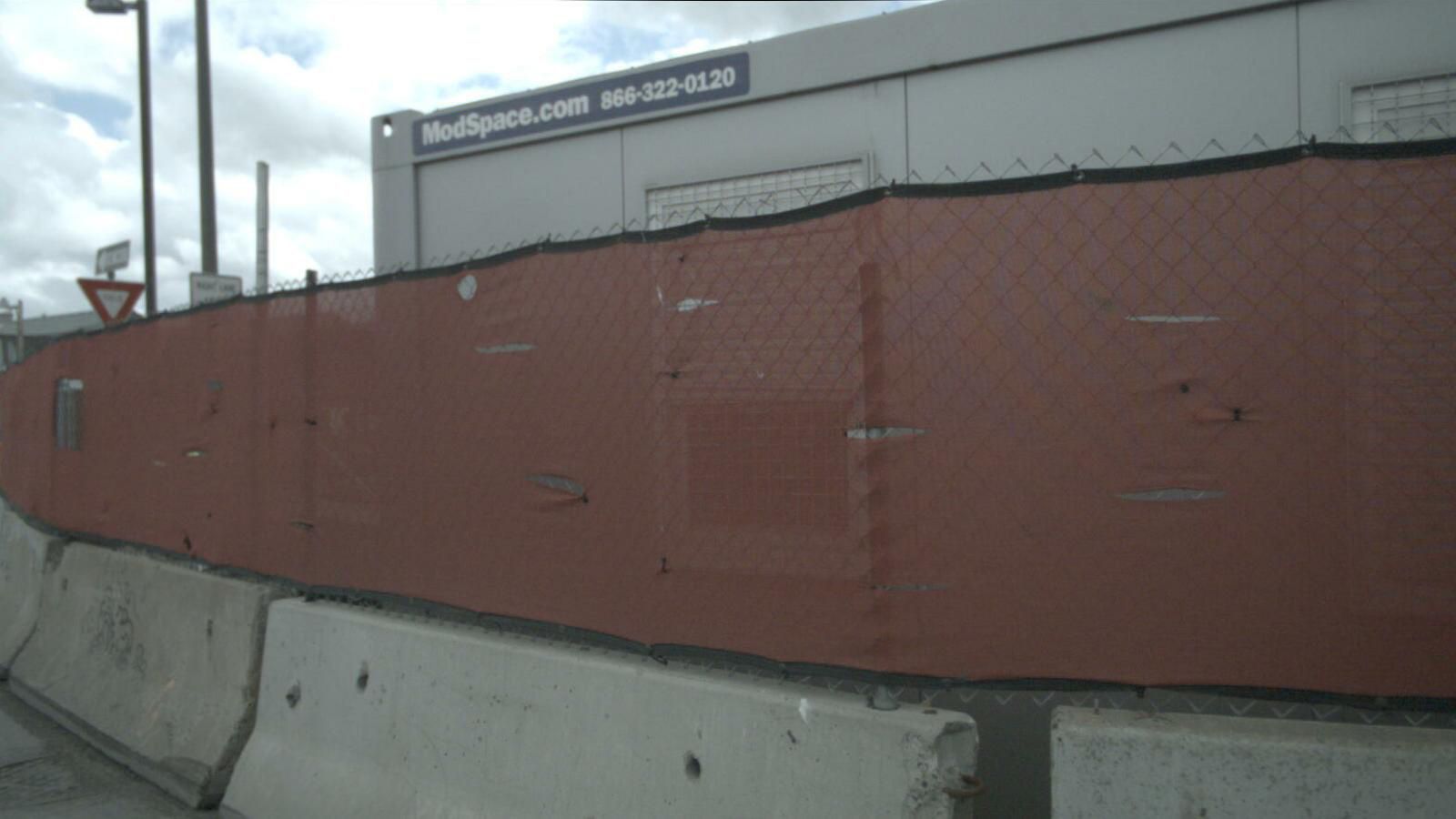}
    \includegraphics[width=\linewidth]{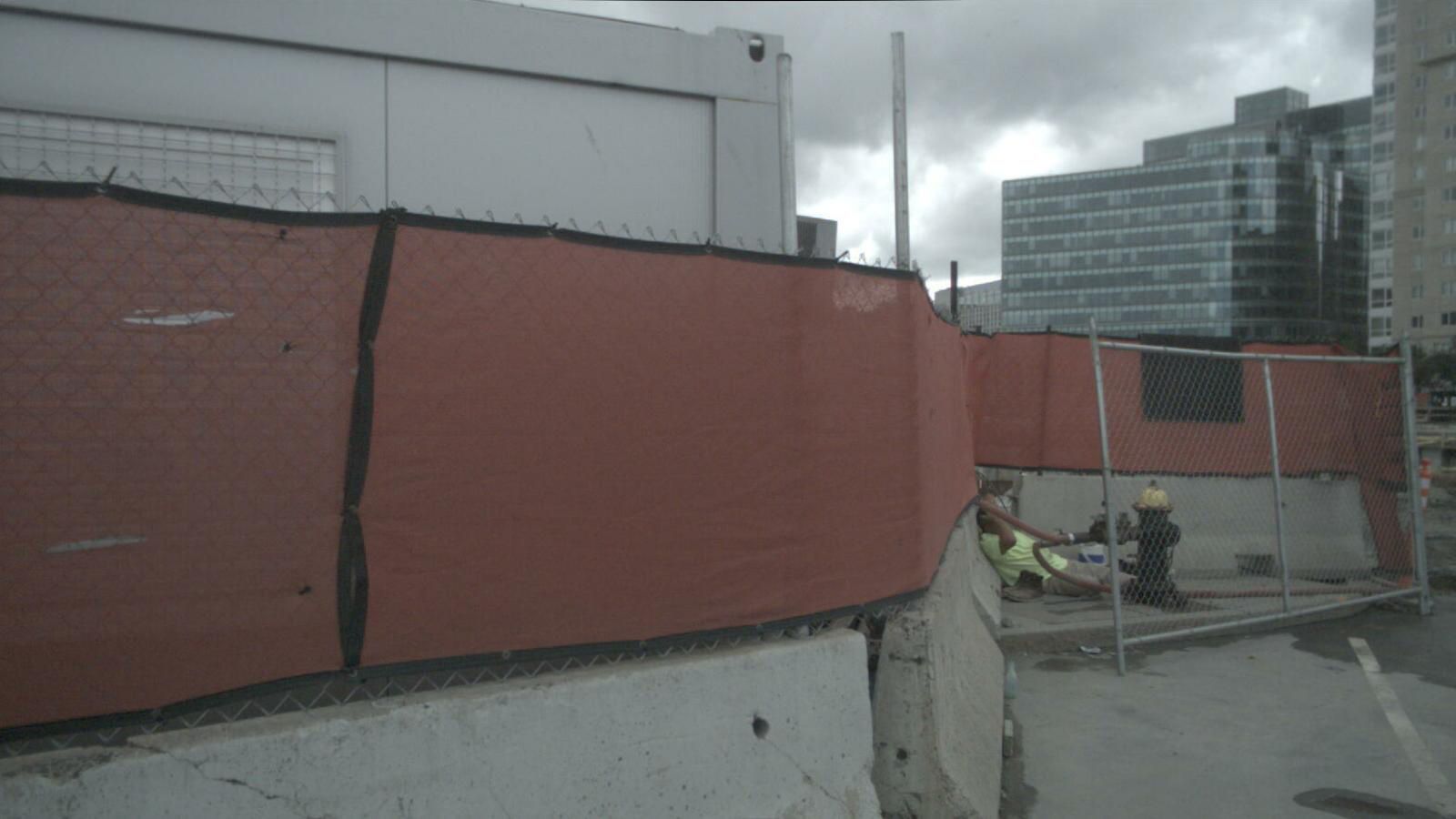}
\end{minipage}%
\hfill %
\begin{minipage}[b]{.74\textwidth}
  \centering
  \begin{subfigure}{.24\linewidth}
    \includegraphics[width=\linewidth]{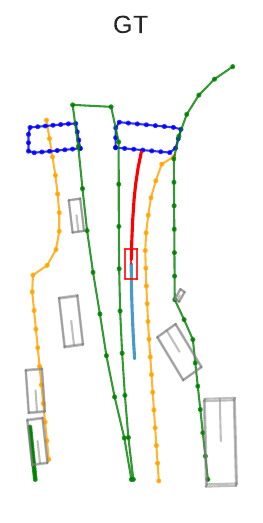}
    \label{fig:vis3_gt}
  \end{subfigure}%
  \begin{subfigure}{.24\linewidth}
    \includegraphics[width=\linewidth]{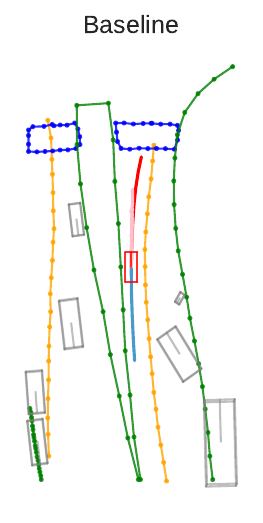}
    \label{fig:vis3_hmn}
  \end{subfigure}%
  \begin{subfigure}{.24\linewidth}
    \includegraphics[width=\linewidth]{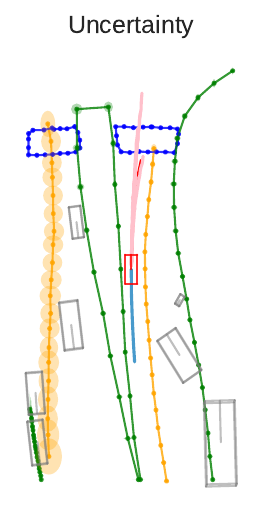}
    \label{fig:vis3_hmu}
  \end{subfigure}%
  \begin{subfigure}{.24\linewidth}
    \includegraphics[width=\linewidth]{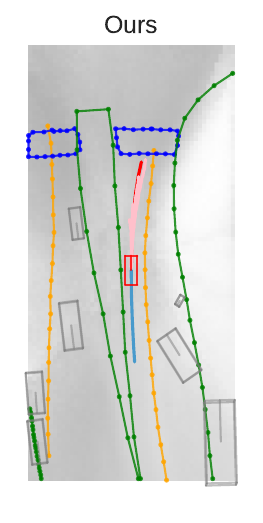}
    \label{fig:vis3_dsu}
  \end{subfigure}

  \vspace{-0.4cm}
  
  \begin{subfigure}{0.92\linewidth}
        \begin{tikzpicture}
        \draw [darkgreen, ultra thick] (0,0.5) -- (0.35,0.5) node[right, black] {\scriptsize Road Boundary};
        \draw [darkyellow, ultra thick] (2.65,0.5) -- (3,0.5) node[right, black] {\scriptsize Lane Divider};
        \draw [blue, ultra thick] (5.0,0.5) -- (5.35,0.5) node[right, black] {\scriptsize Pedestrian Crossing};
        \draw [red, ultra thick] (0,0) -- (0.35,0) node[right, black] {\scriptsize GT Future};
        \draw [pink, ultra thick] (2.1,0) -- (2.45,0) node[right, black] {\scriptsize Predicted Trajectories};
        \draw [historyblue, ultra thick] (5.65,0) -- (6,0) node[right, black] {\scriptsize Agent History};
        \end{tikzpicture}
  \end{subfigure}
\end{minipage}

\vspace{-0.2cm}

\caption{StreamMapNet~\cite{yuan2024streammapnet} and DenseTNT~\cite{GuSunEtAl2021} combined using the strategy in \cref{sec:strategy_3}. By replacing agent trajectory information with BEV features, DenseTNT is able to predict trajectories that stop before the crosswalk, compared to the undershooting and overshooting of the Baseline and Uncertainty-enhanced~\cite{GuSongEtAl2024} approaches. }
\label{fig:vis3}

\vspace{-0.5cm}

\end{figure*}

%% file: sec/5_conclusion.tex
\section{Conclusion}
\vspace{-0.2cm}
\label{sec:conclusion}
In this work, we propose three different strategies to leverage the intermediate BEV features within online map estimation models in downstream tasks such as behavior prediction. We systematically evaluate the benefits of different BEV encoding strategies and demonstrate how incorporating BEV features in downstream behavior prediction results in significant performance and runtime improvements. In particular, combinations of various online mapping and prediction methods achieve up to $73\%$ faster inference times when operating directly from intermediate BEV features and produce predictions that are up to $29\%$ more accurate across a variety of evaluation metrics.

Our work's limitations and potential negative impacts relate to its use of black-box features in lieu of vectorized map estimation. While this yields performance and runtime improvements, it may complicate introspection into why a behavior prediction algorithm made certain predictions (compared to when explicit map elements are encoded). Towards this end, exciting future directions include further explorations of mapping models' BEV feature spaces, strategies to interpret BEV features at runtime (alternatives to costly decoding processes), and co-training strategies to inform upstream map estimation models of the task of behavior prediction (ideally yielding improvements in both mapping and prediction performance, towards the development of end-to-end AV stacks).

%% file: sec/6_suppl.tex
\clearpage
\appendix

\section{Training Details}
\label{sec:supp_training}
\vspace{-0.5cm}
\begin{table}
\centering
\resizebox{\linewidth}{!}{
  \begin{tabular}{@{}l|cc|ccccc@{}}
    \toprule
    Stage & \multicolumn{2}{c|}{Prediction} & \multicolumn{5}{c}{BEV Feature Attention} \\
    \midrule
    Map Model & LR & Weight Decay & Patch Size & $\text{MLP}_{\text{dim}}$ & Depth & Heads & $\text{Head}_{\text{dim}}$ \\
    \midrule
    MapTR~\cite{MapTR} & $5\text{E-}4$ & $1\text{E-}4$ & $(20, 10)$ & $512$ & $6$ & $16$ & $64$ \\
    MapTRv2~\cite{maptrv2} & $3.5\text{E-}4$ & $1\text{E-}2$ & $(20, 10)$ & $64$ & $6$ & $16$ & $64$ \\
    MapTRv2-Centerline~\cite{maptrv2} & $3.5\text{E-}4$ & $1\text{E-}2$ & $(20, 20)$ & $64$ & $4$ & $12$ & $32$ \\
    StreamMapNet~\cite{yuan2024streammapnet} & $3.5\text{E-}4$ & $1\text{E-}3$ & $(10, 5)$ & $128$ & $6$ & $16$ & $64$ \\
    \bottomrule
  \end{tabular}
}
  \vspace{0.1cm}
  \caption{The hyperparameters used when training HiVT~\cite{zhou2022hivt} with various online mapping models in \cref{sec:replace_attend}, where agent-lane attention is replaced with agent-BEV attention.}
  \label{tab:supp_hyperparams_1}

\vspace{-1.5cm}
  
\end{table}

\begin{table}
  \centering
  \begin{tabular}{@{}l|ccc@{}}
    \toprule
    Map Model & LR & Weight Decay & Dropout \\
    \midrule
    MapTR~\cite{MapTR} & $1.5\text{E-}4$ & $0.05$ & $0.2$  \\
    MapTRv2~\cite{maptrv2} & $1.5\text{E-}4$ & $0.05$ & $0.2$  \\
    MapTRv2-Centerline~\cite{maptrv2} & $2\text{E-}4$ & $0.05$ & $0.2$ \\
    \bottomrule
  \end{tabular}
  \vspace{0.1cm}
  \caption{The hyperparameters used when training DenseTNT~\cite{GuSunEtAl2021} with various online mapping models in \cref{sec:strategy_2}, where lane vectors are enhanced with BEV grid features.}
  \label{tab:supp_hyperparams_2}

\vspace{-1.5cm}

\end{table}

\begin{table}
\centering
  \begin{tabular}{@{}cc|ccccc@{}}
    \toprule
    \multicolumn{2}{c|}{Prediction} & \multicolumn{5}{c}{BEV Feature Attention} \\
    \midrule
    LR & Weight Decay & Patch Size & $\text{MLP}_{\text{dim}}$ & Depth & Heads & $\text{Head}_{\text{dim}}$ \\
    \midrule
    $5\text{E-}4$ & $1\text{E-}2$ & $(10, 5)$ & $128$ & $6$ & $16$ & $64$ \\
    \bottomrule
  \end{tabular}
  \vspace{0.1cm}
  \caption{The hyperparameters used when training DenseTNT~\cite{GuSunEtAl2021} with StreamMapNet~\cite{yuan2024streammapnet} in \cref{sec:strategy_3}, where agent information is replaced with temporal BEV feature attention.}
  \label{tab:supp_hyperparams_3}

\vspace{-0.5cm}
  
\end{table}

\subsection{Data Preprocessing}
To ensure a fair comparison across different map estimation and prediction models, we unify the orientations of the BEV features and the resulting estimated map. The scene is centered at the ego-vehicle frame, with the positive y-axis aligned with the forward-moving direction, and the positive x-axis aligned with the right side of the ego-vehicle. The BEV features are adjusted accordingly. The perception range ($H \times W$) is $60 m \times 30 m$. Due to the limits of AV perception, we only predict for agents within this perception range.

\subsection{Model Training}
To address the potential variability in convergence rates between different integration approaches and map-prediction combinations, each model is individually adjusted to optimize performance. The BEV dimension is $200\times100$ for MapTR models~\cite{MapTR,maptrv2} and $100\times50$ for StreamMapNet~\cite{yuan2024streammapnet}. 

During training, online map estimation models are trained first as in~\cite{GuSongEtAl2024}. This produces map element polylines with corresponding uncertainties. During inference, we also save the BEV features produced by the trained models, providing the necessary data for all three settings for prediction: Baseline, where only lane vectors are used; Uncertainty, where uncertainty is incorporated; and our approach, where BEV features are incorporated. After we obtain this modified dataset, HiVT~\cite{zhou2022hivt} and DenseTNT~\cite{GuSunEtAl2021} are trained following the different strategies in \cref{sec:methods}. 

In \cref{sec:replace_attend}, HiVT's local encoder is modified by replacing agent-lane features with agent-BEV features. As seen in \cref{tab:supp_hyperparams_1}, we reduce the attention module size as the complexity of the mapping model increases. This adjustment is shown via the decrease in MLP layer size and head dimension across the MapTR series. The increase in BEV patch size for MapTRv2-Centerline compared to MapTRv2 also indicates a coarser feature representation. The output dimension of the attention module is adjusted to match the original agent-lane feature dimension, ensuring compatibility with the HiVT's global interaction module. 

For the approach in \cref{sec:strategy_2}, we tune the prediction training hyperparameters to accommodate the additional information provided by BEV features, as seen in \cref{tab:supp_hyperparams_2}. Due to the increased complexity of input data, the learning rate is reduced to the order of $10^{-4}$ and weight decay is increased to 0.05 from 0.01 to ensure smooth training convergence. Dropout is also increased slightly from $0.1$ to $0.2$. When encoding lane information in the point-level subgraph of Vectornet, the hidden layer size is doubled to accommodate the extra BEV features after concatenating them with the original raw lane vertices. 

The hyperparameter choices for \cref{sec:strategy_3} are shown in \cref{tab:supp_hyperparams_3}. Prediction model values are adjusted in the same way as \cref{tab:supp_hyperparams_2}, with a smaller learning rate to ensure convergence. For the BEV attention module, the hyperparameter choices are the same as in the corresponding row of \cref{tab:supp_hyperparams_1}. 

\section{Additional Quantitative Comparisons}

\subsection{Runtime Comparisons}

Below, runtime is measured on an RTX 4090 GPU from when raw RGB camera images are input to when trajectories are produced.

\begin{table}[h]
\centering
\footnotesize
\begin{tabular}{l|c|c}
\toprule
\textbf{Model Combination} & \textbf{Base (ms)} & \textbf{Ours (ms)} \\ \midrule
HiVT + MapTR & 22.4 & 9.1 \\ 
HiVT + MapTRv2 & 26.7 & 13 \\ 
HiVT + StreamMapNet & 33.6 & 29.4 \\
\bottomrule
\end{tabular}
\end{table}

\section{Additional Visualizations}

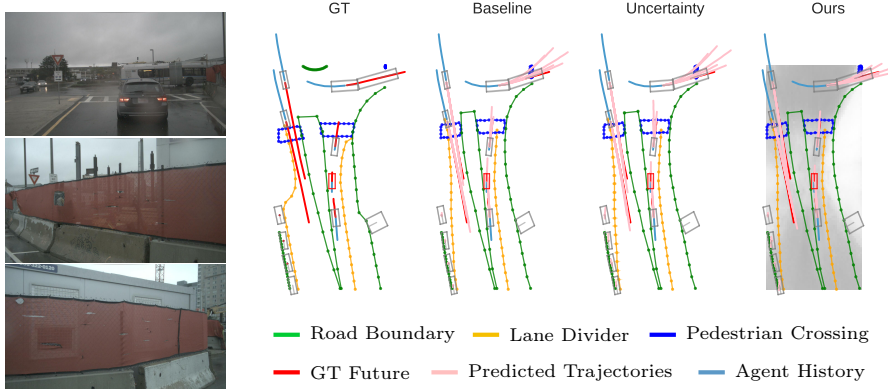
\begin{figure*}
\centering

\begin{minipage}[b]{.24\textwidth}
    \includegraphics[width=\linewidth]{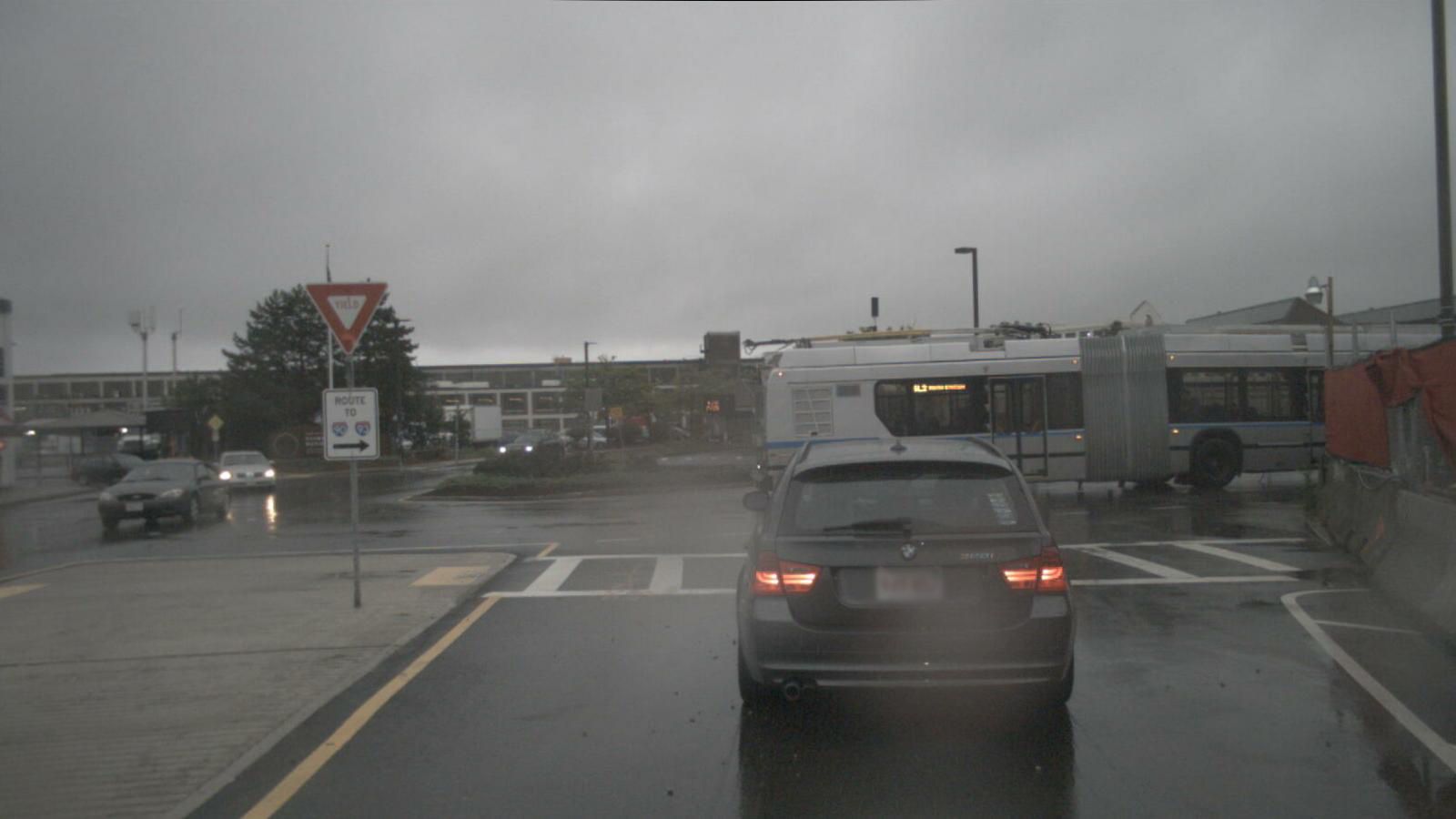}
    \includegraphics[width=\linewidth]{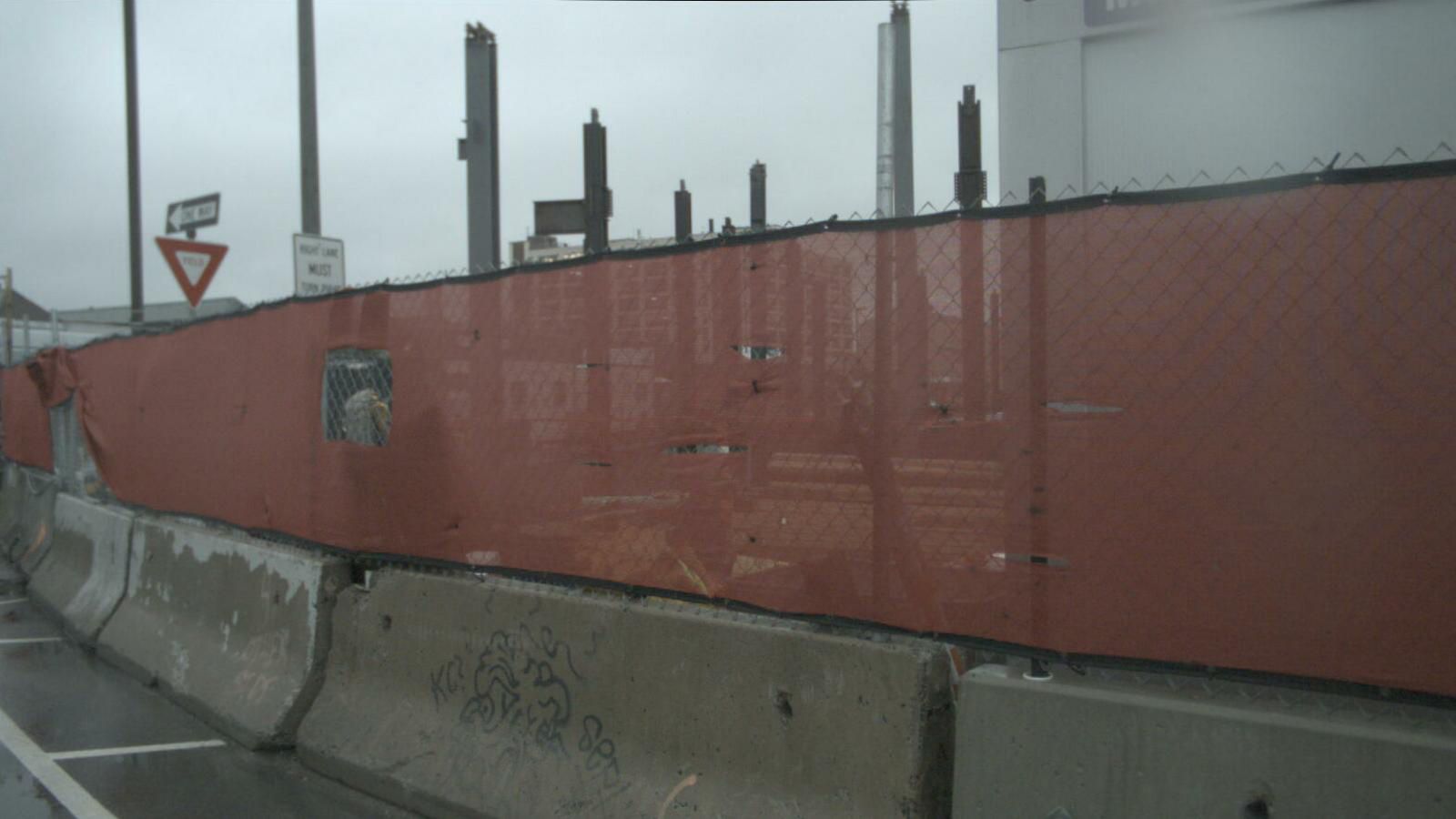}
    \includegraphics[width=\linewidth]{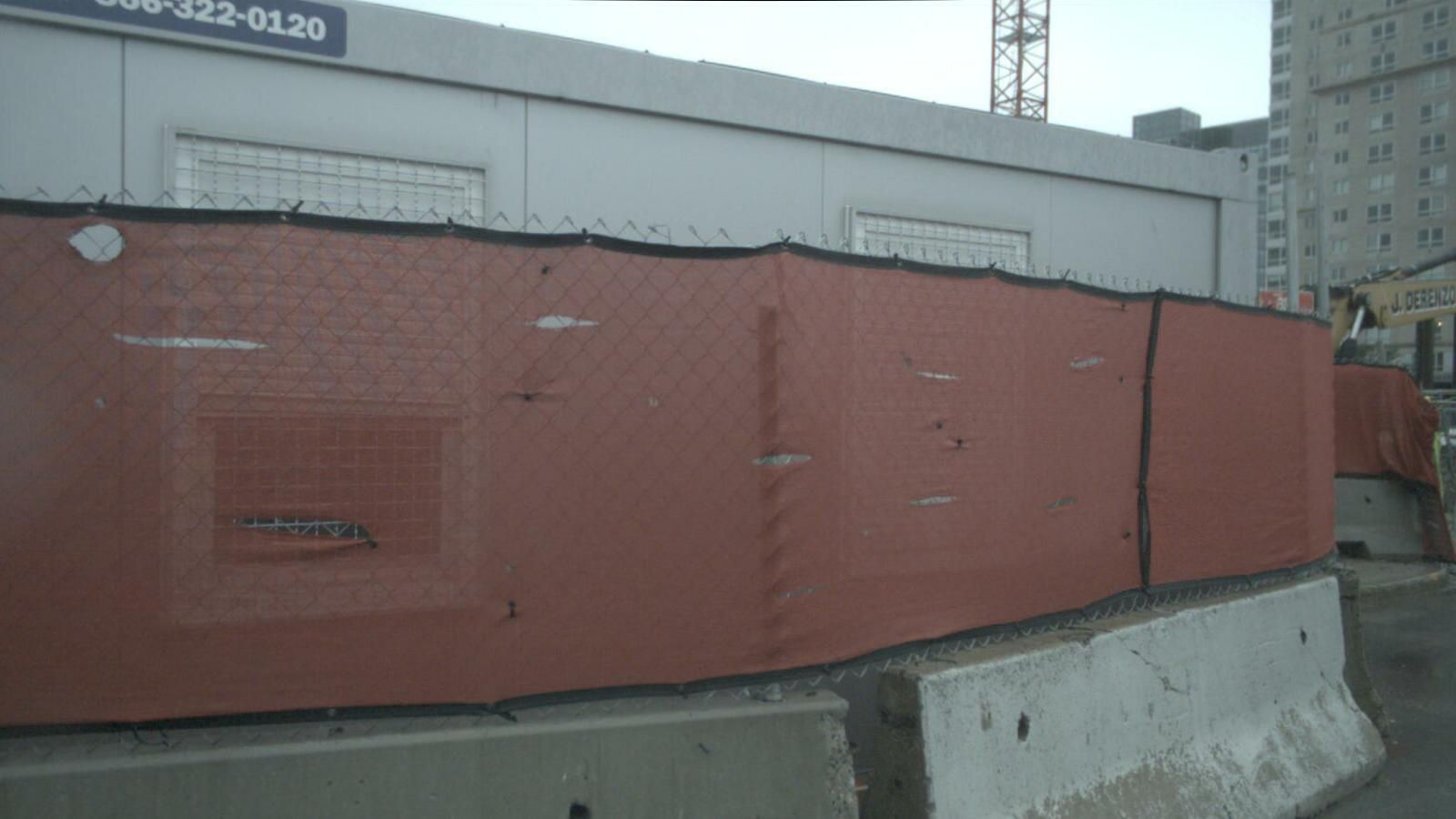}
\end{minipage}%
\hfill %
\begin{minipage}[b]{.74\textwidth}
  \centering
  \begin{subfigure}{.24\linewidth}
    \includegraphics[width=\linewidth]{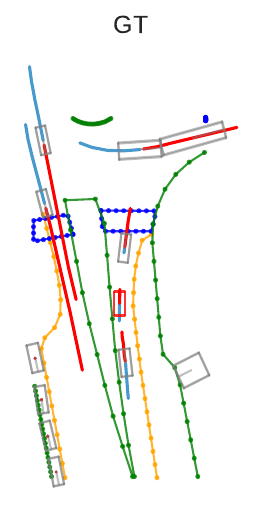}
    \label{fig:vis4_gt}
  \end{subfigure}%
  \begin{subfigure}{.24\linewidth}
    \includegraphics[width=\linewidth]{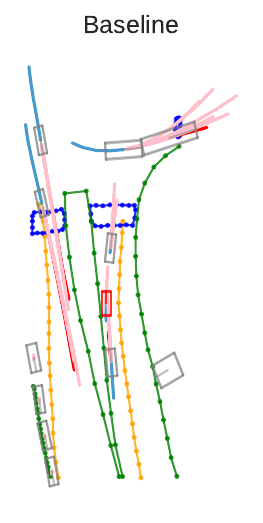}
    \label{fig:vis4_hmn}
  \end{subfigure}%
  \begin{subfigure}{.24\linewidth}
    \includegraphics[width=\linewidth]{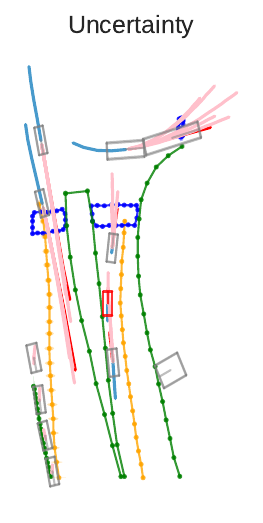}
    \label{fig:vis4_hmu}
  \end{subfigure}%
  \begin{subfigure}{.24\linewidth}
    \includegraphics[width=\linewidth]{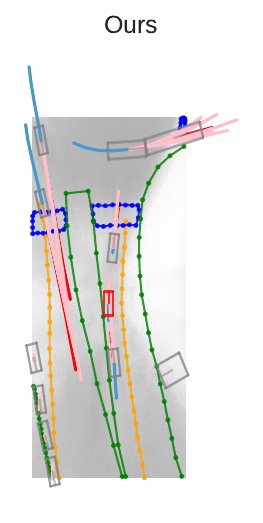}
    \label{fig:vis4_dsu}
  \end{subfigure}

    \vspace{-0.4cm}

    \begin{subfigure}{0.92\linewidth}
        \begin{tikzpicture}
        \draw [darkgreen, ultra thick] (0,0.5) -- (0.35,0.5) node[right, black] {\scriptsize Road Boundary};
        \draw [darkyellow, ultra thick] (2.65,0.5) -- (3,0.5) node[right, black] {\scriptsize Lane Divider};
        \draw [blue, ultra thick] (5.0,0.5) -- (5.35,0.5) node[right, black] {\scriptsize Pedestrian Crossing};
        \draw [red, ultra thick] (0,0) -- (0.35,0) node[right, black] {\scriptsize GT Future};
        \draw [pink, ultra thick] (2.1,0) -- (2.45,0) node[right, black] {\scriptsize Predicted Trajectories};
        \draw [historyblue, ultra thick] (5.65,0) -- (6,0) node[right, black] {\scriptsize Agent History};
        \end{tikzpicture}
  \end{subfigure}

\end{minipage}

\vspace{-0.2cm}

\caption{StreamMapNet~\cite{yuan2024streammapnet} and HiVT~\cite{zhou2022hivt} combined using the strategy in \cref{sec:replace_attend}. By replacing lane information with temporal BEV features, HiVT is able to better predict stopping behavior, avoiding overshooting the GT (as in the Baseline and Uncertainty-enhanced approach).}
\label{fig:vis4}

\vspace{-0.5cm}

\end{figure*}

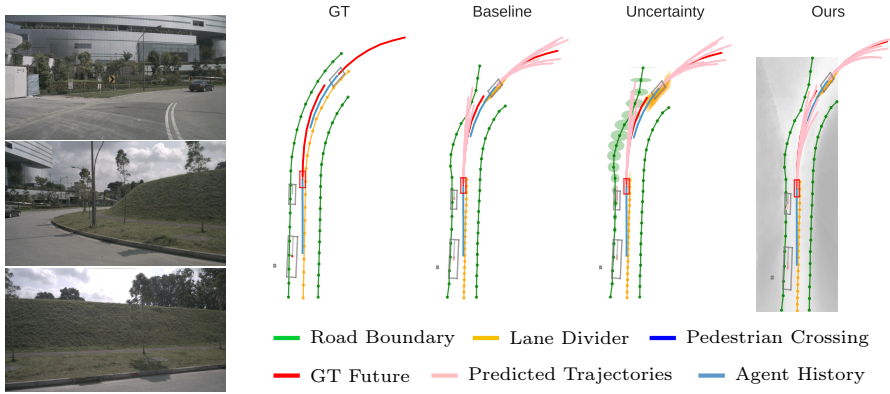
\begin{figure*}[t]
\centering

\begin{minipage}[b]{.24\textwidth}
    \includegraphics[width=\linewidth]{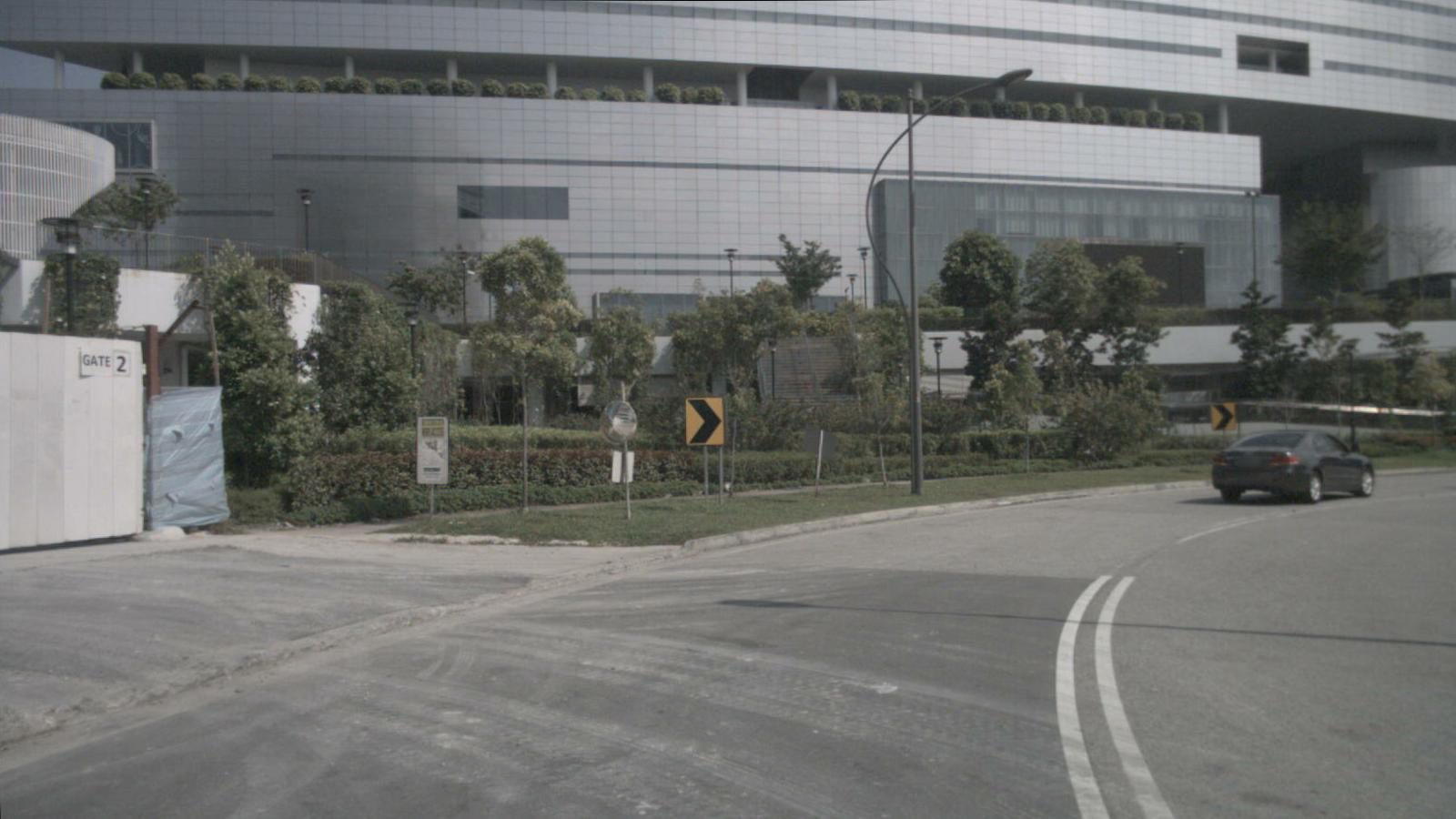}
    \includegraphics[width=\linewidth]{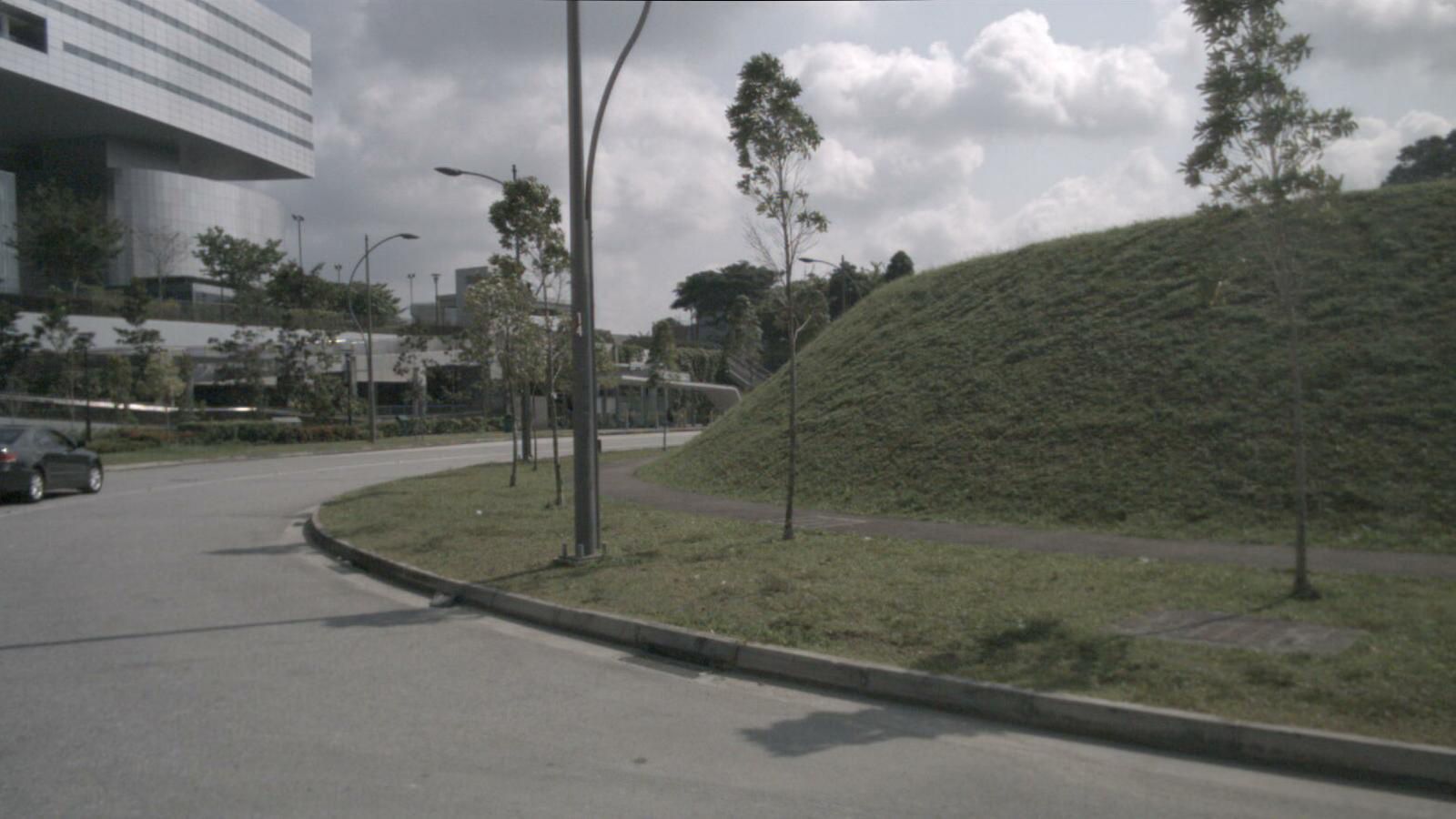}
    \includegraphics[width=\linewidth]{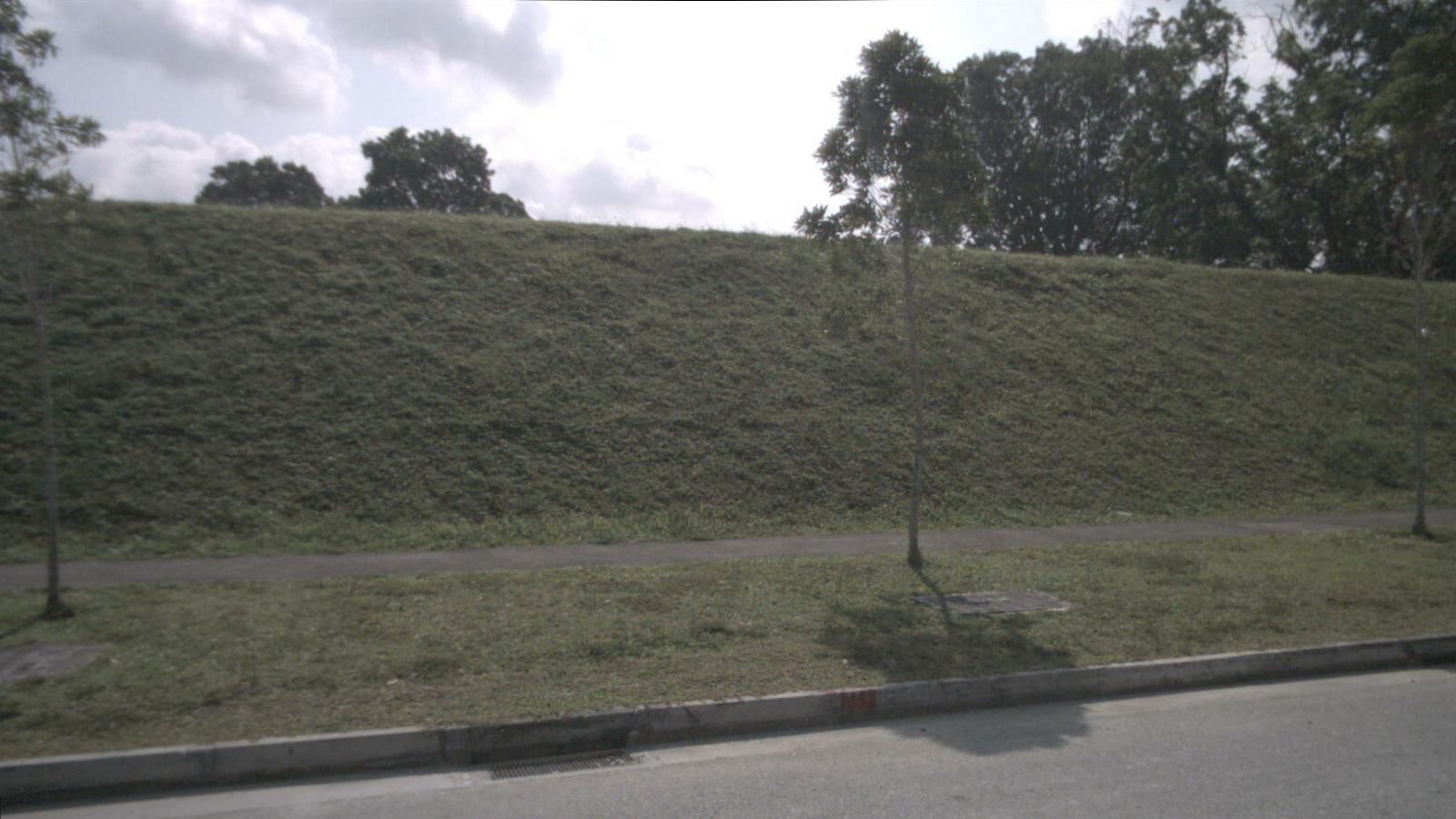}
\end{minipage}%
\hfill %
\begin{minipage}[b]{.74\textwidth}
  \centering
  \begin{subfigure}{.24\linewidth}
    \includegraphics[width=\linewidth]{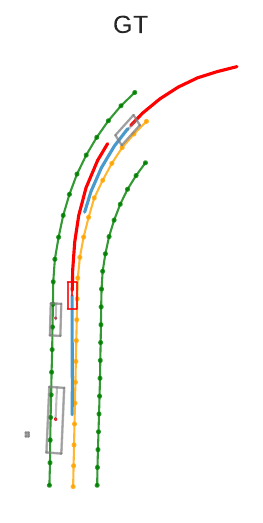}
    \label{fig:vis5_gt}
  \end{subfigure}%
  \begin{subfigure}{.24\linewidth}
    \includegraphics[width=\linewidth]{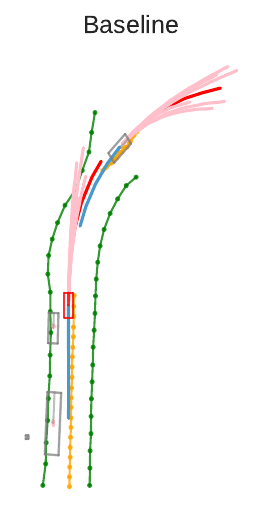}
    \label{fig:vis5_hmn}
  \end{subfigure}%
  \begin{subfigure}{.24\linewidth}
    \includegraphics[width=\linewidth]{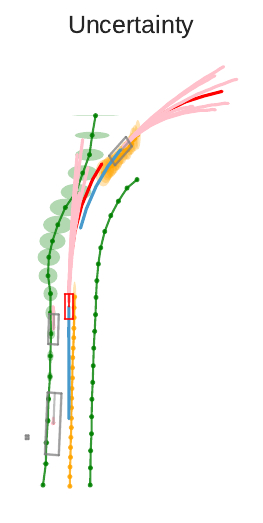}
    \label{fig:vis5_hmu}
  \end{subfigure}%
  \begin{subfigure}{.24\linewidth}
    \includegraphics[width=\linewidth]{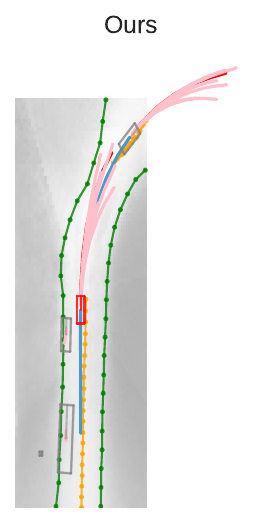}
    \label{fig:vis5_dsu}
  \end{subfigure}

    \vspace{-0.4cm}

    \begin{subfigure}{0.92\linewidth}
        \begin{tikzpicture}
        \draw [darkgreen, ultra thick] (0,0.5) -- (0.35,0.5) node[right, black] {\scriptsize Road Boundary};
        \draw [darkyellow, ultra thick] (2.65,0.5) -- (3,0.5) node[right, black] {\scriptsize Lane Divider};
        \draw [blue, ultra thick] (5.0,0.5) -- (5.35,0.5) node[right, black] {\scriptsize Pedestrian Crossing};
        \draw [red, ultra thick] (0,0) -- (0.35,0) node[right, black] {\scriptsize GT Future};
        \draw [pink, ultra thick] (2.1,0) -- (2.45,0) node[right, black] {\scriptsize Predicted Trajectories};
        \draw [historyblue, ultra thick] (5.65,0) -- (6,0) node[right, black] {\scriptsize Agent History};
        \end{tikzpicture}
  \end{subfigure}

\end{minipage}

\vspace{-0.2cm}

\caption{StreamMapNet~\cite{yuan2024streammapnet} and HiVT~\cite{zhou2022hivt} combined using the strategy in \cref{sec:replace_attend}. By replacing lane information with temporal BEV features, HiVT's predictions respect boundaries, in contrast to both the Baseline and Uncertainty-enhanced approaches which deviate outside the green road boundary. Further, our approach's predicted trajectories align more closely to the GT.}
\label{fig:vis5}

\vspace{-0.6cm}

\end{figure*}

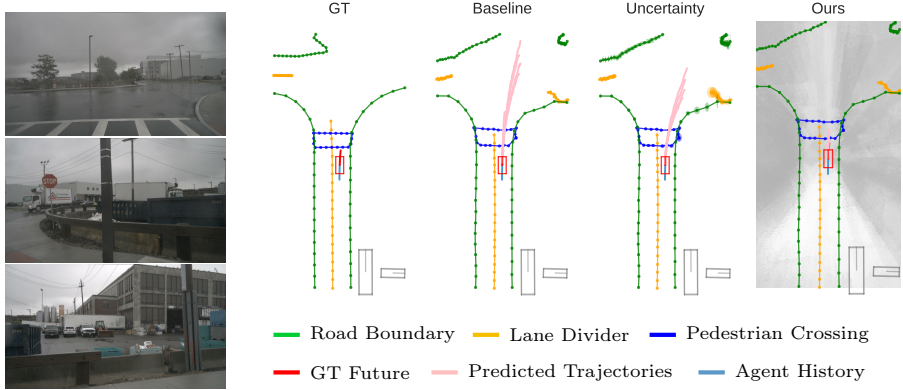
\begin{figure*}[t]
\centering

\begin{minipage}[b]{.24\textwidth}
    \includegraphics[width=\linewidth]{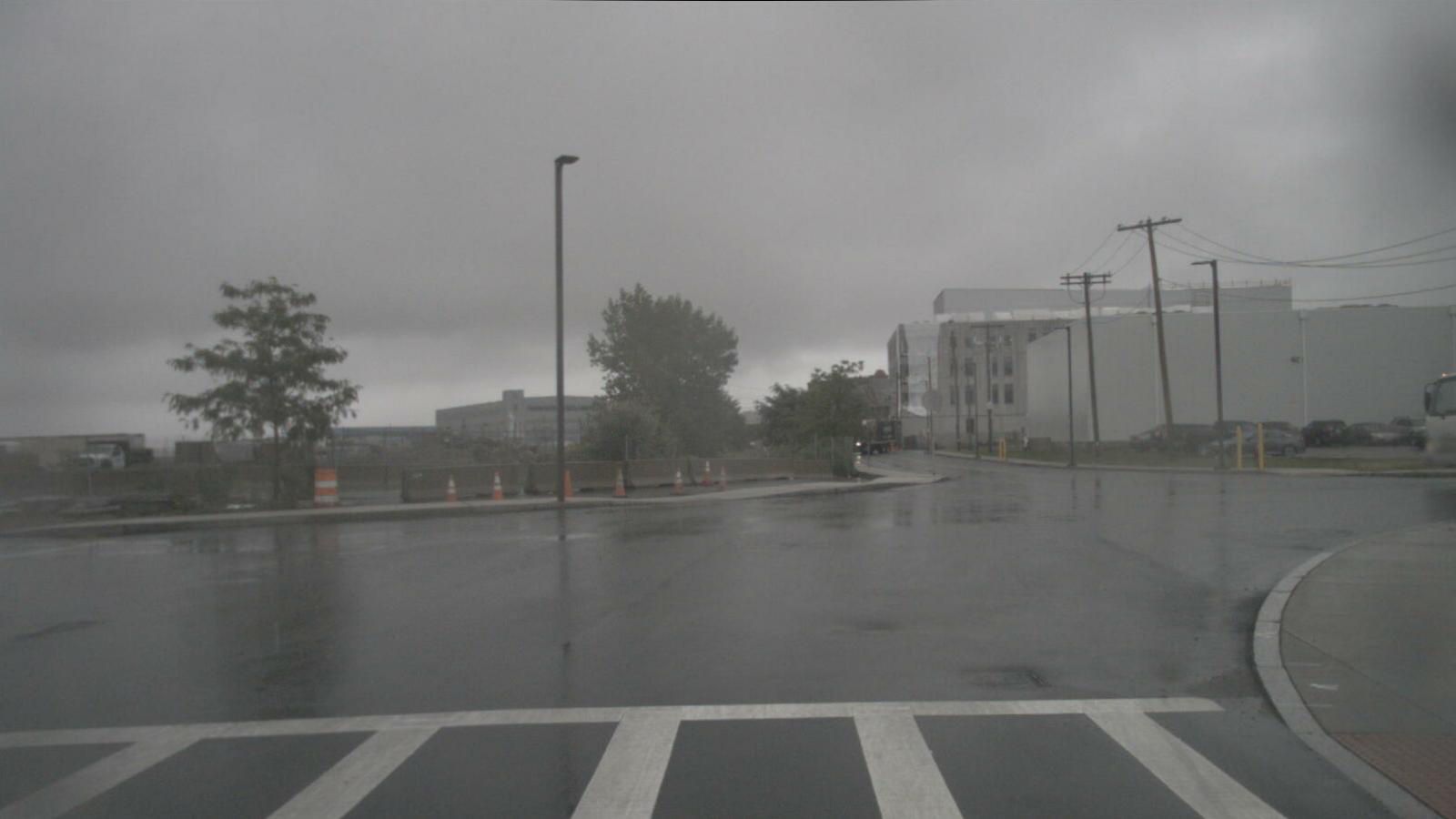}
    \includegraphics[width=\linewidth]{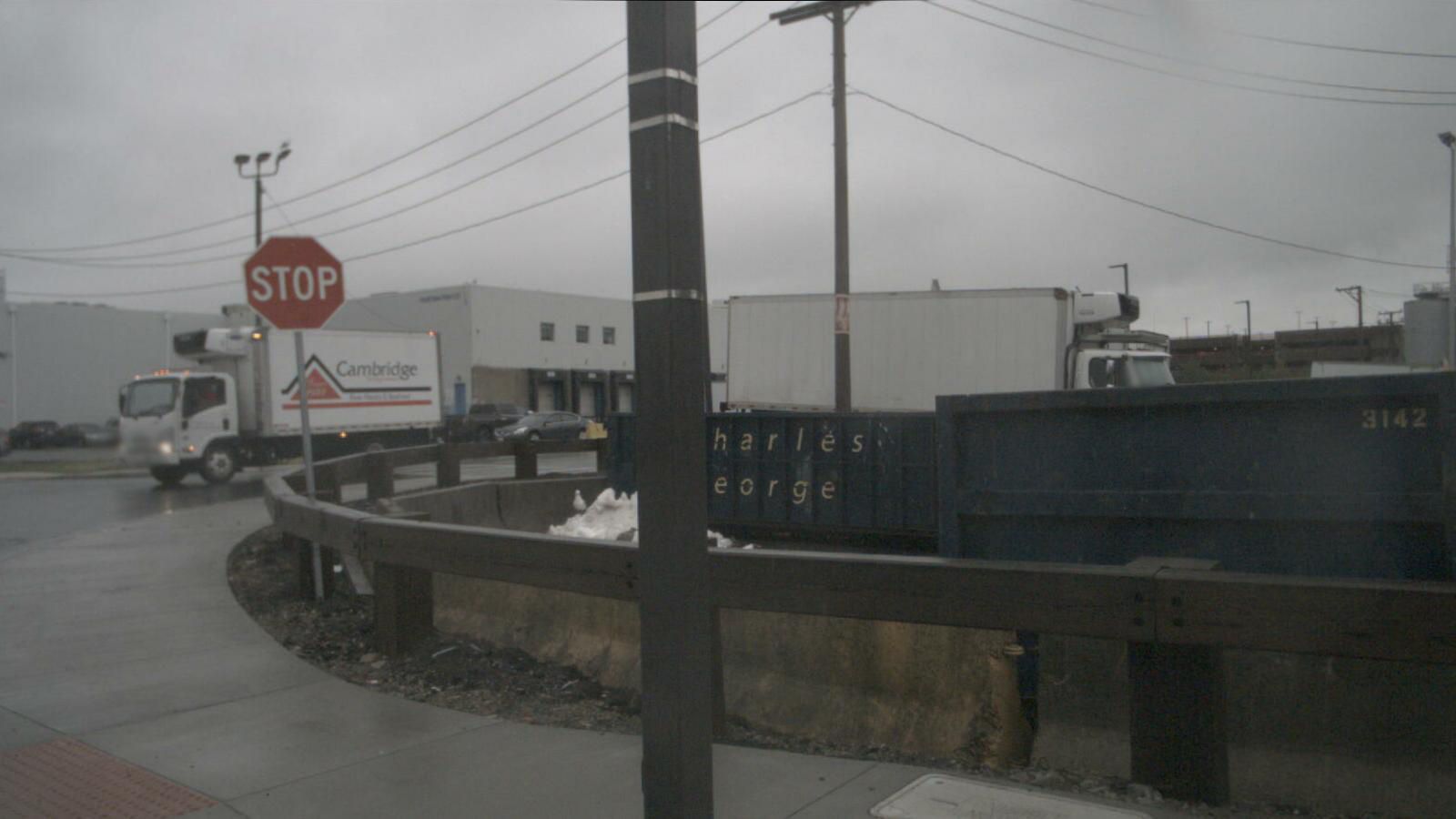}
    \includegraphics[width=\linewidth]{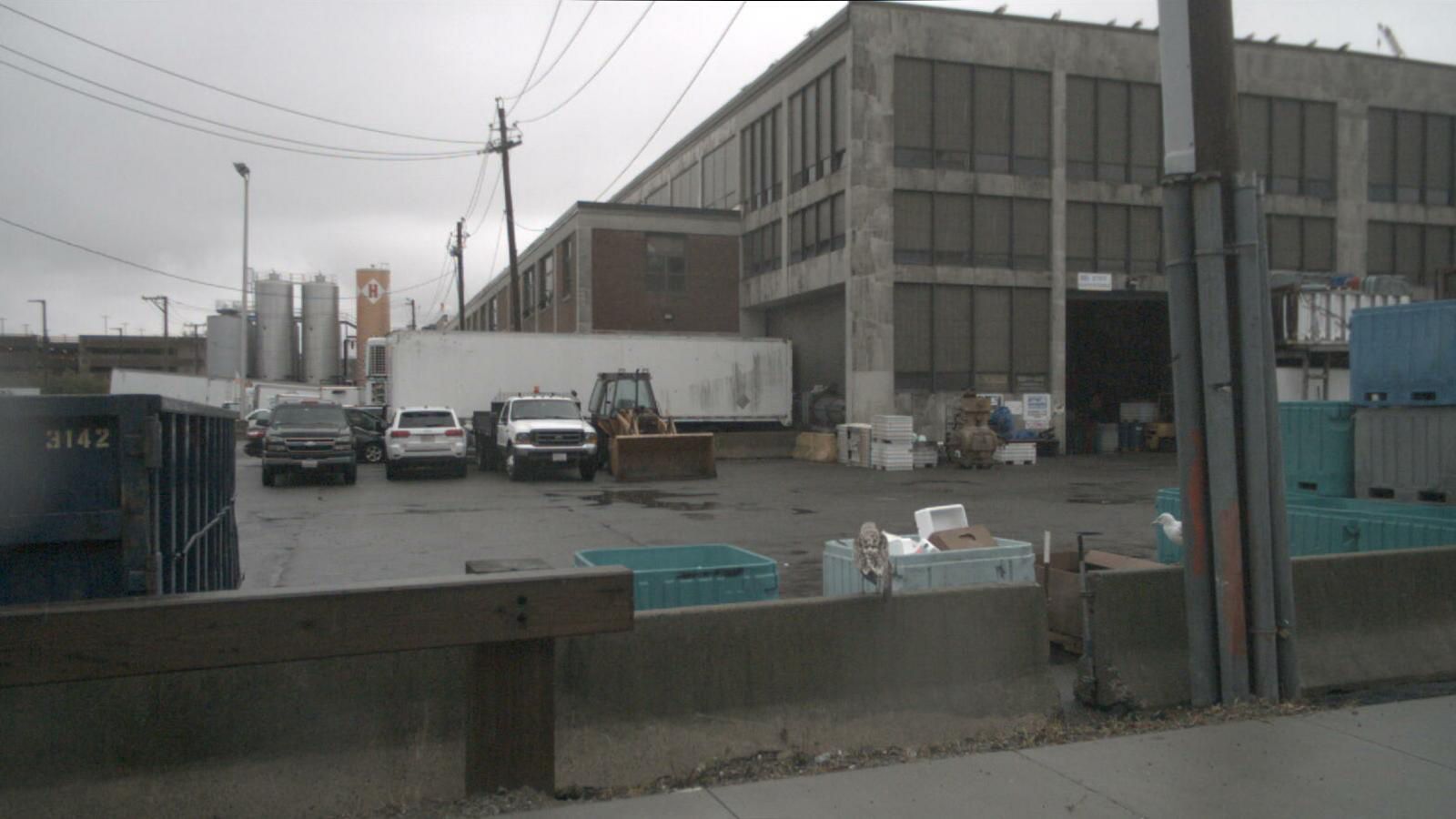}
\end{minipage}%
\hfill %
\begin{minipage}[b]{.74\textwidth}
  \centering
  \begin{subfigure}{.24\linewidth}
    \includegraphics[width=\linewidth]{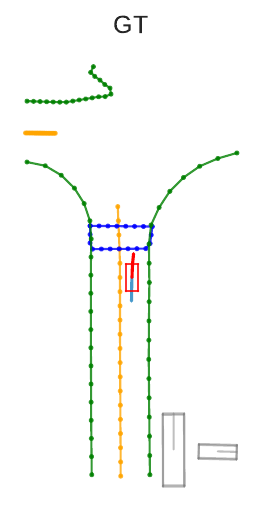}
    \label{fig:vis6_gt}
  \end{subfigure}%
  \begin{subfigure}{.24\linewidth}
    \includegraphics[width=\linewidth]{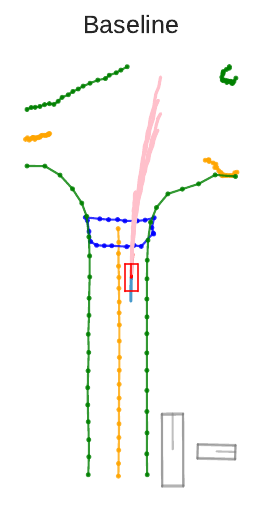}
    \label{fig:vis6_hmn}
  \end{subfigure}%
  \begin{subfigure}{.24\linewidth}
    \includegraphics[width=\linewidth]{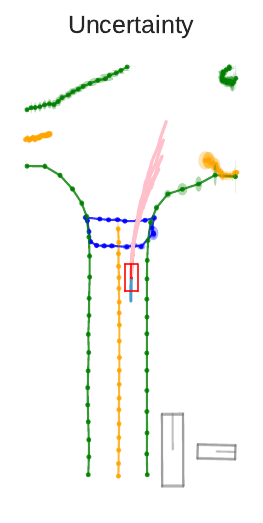}
    \label{fig:vis6_hmu}
  \end{subfigure}%
  \begin{subfigure}{.24\linewidth}
    \includegraphics[width=\linewidth]{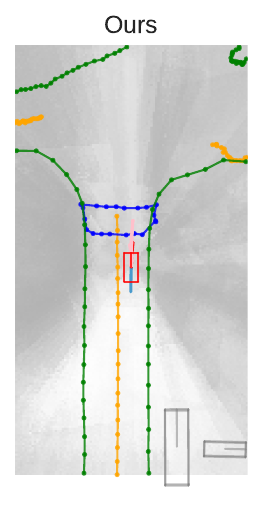}
    \label{fig:vis6_dsu}
  \end{subfigure}

    \vspace{-0.4cm}

    \begin{subfigure}{0.92\linewidth}
        \begin{tikzpicture}
        \draw [darkgreen, ultra thick] (0,0.5) -- (0.35,0.5) node[right, black] {\scriptsize Road Boundary};
        \draw [darkyellow, ultra thick] (2.65,0.5) -- (3,0.5) node[right, black] {\scriptsize Lane Divider};
        \draw [blue, ultra thick] (5.0,0.5) -- (5.35,0.5) node[right, black] {\scriptsize Pedestrian Crossing};
        \draw [red, ultra thick] (0,0) -- (0.35,0) node[right, black] {\scriptsize GT Future};
        \draw [pink, ultra thick] (2.1,0) -- (2.45,0) node[right, black] {\scriptsize Predicted Trajectories};
        \draw [historyblue, ultra thick] (5.65,0) -- (6,0) node[right, black] {\scriptsize Agent History};
        \end{tikzpicture}
  \end{subfigure}

\end{minipage}

\vspace{-0.2cm}

\caption{MapTR~\cite{MapTR} and DenseTNT~\cite{GuSunEtAl2021} combined via the strategy in \cref{sec:strategy_2}. Our augmentation of map vertices with BEV features enables DenseTNT to produce accurate trajectories, preventing overshooting at an intersection as seen in the Baseline and Uncertainty-enhanced setups.}
\label{fig:vis6}

\vspace{-0.6cm}

\end{figure*}

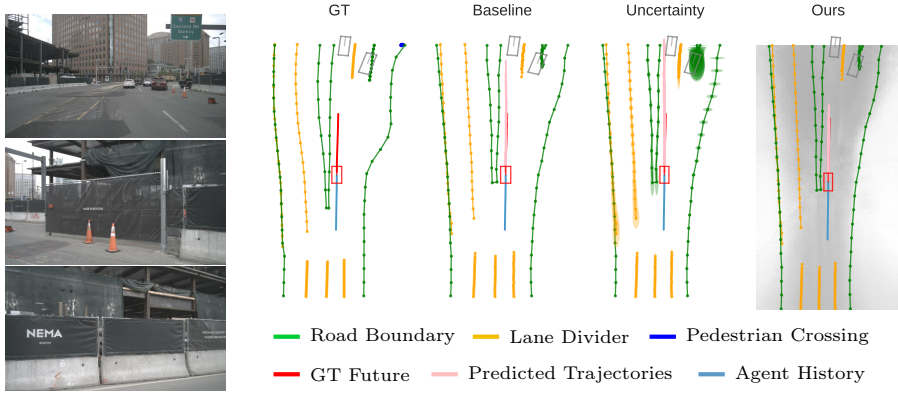
\begin{figure*}[t]
\centering

\begin{minipage}[b]{.24\textwidth}
    \includegraphics[width=\linewidth]{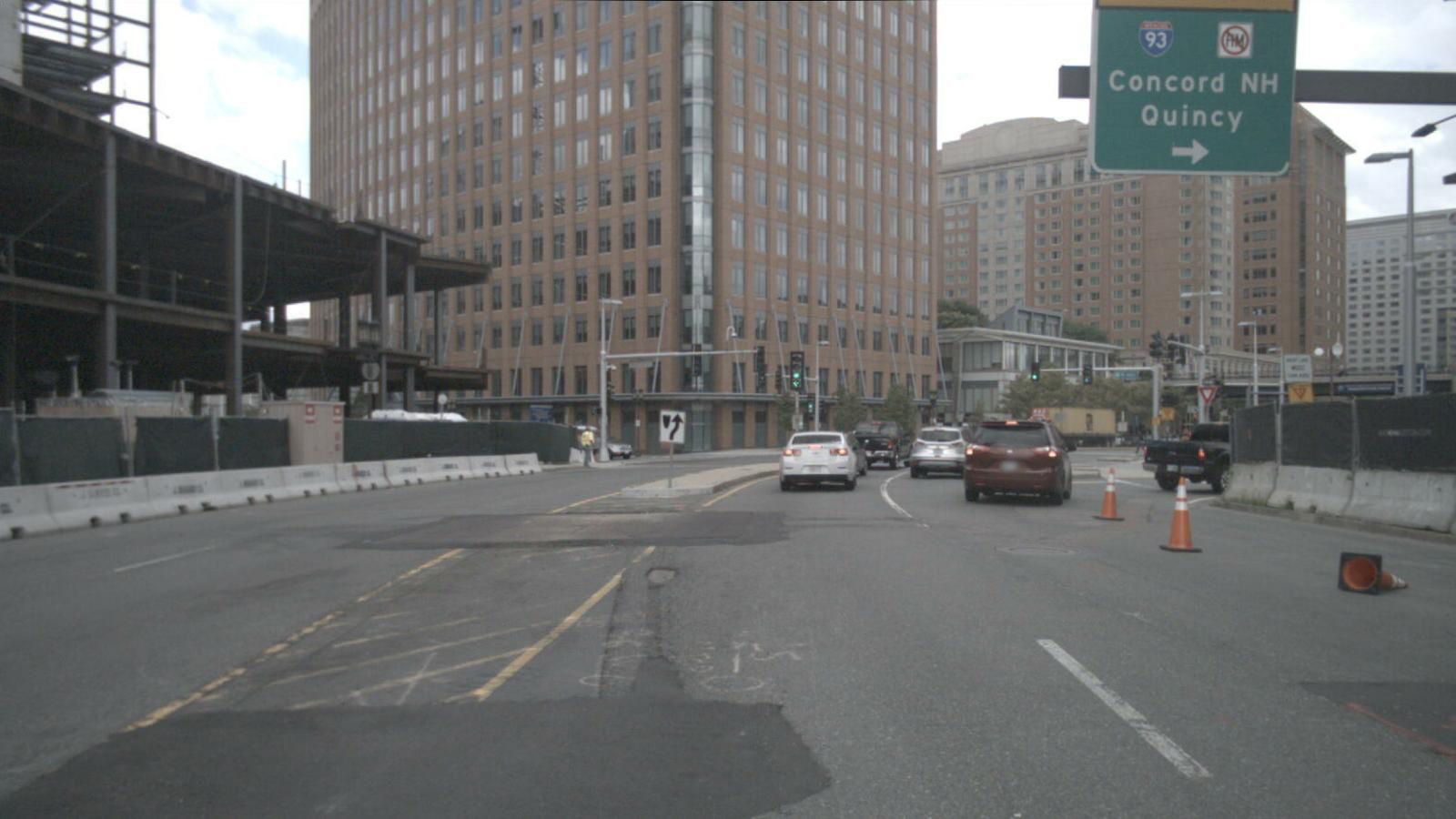}
    \includegraphics[width=\linewidth]{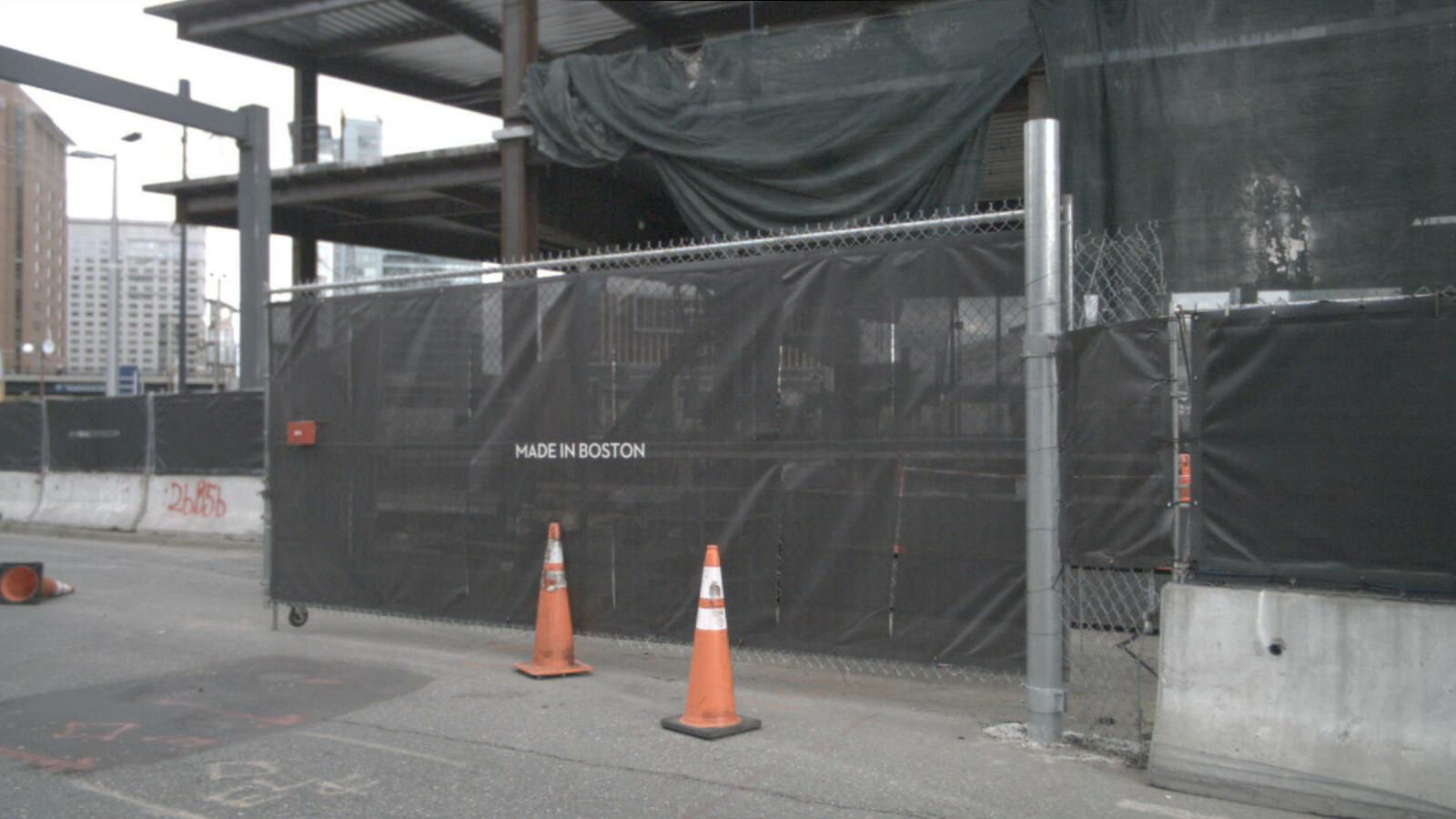}
    \includegraphics[width=\linewidth]{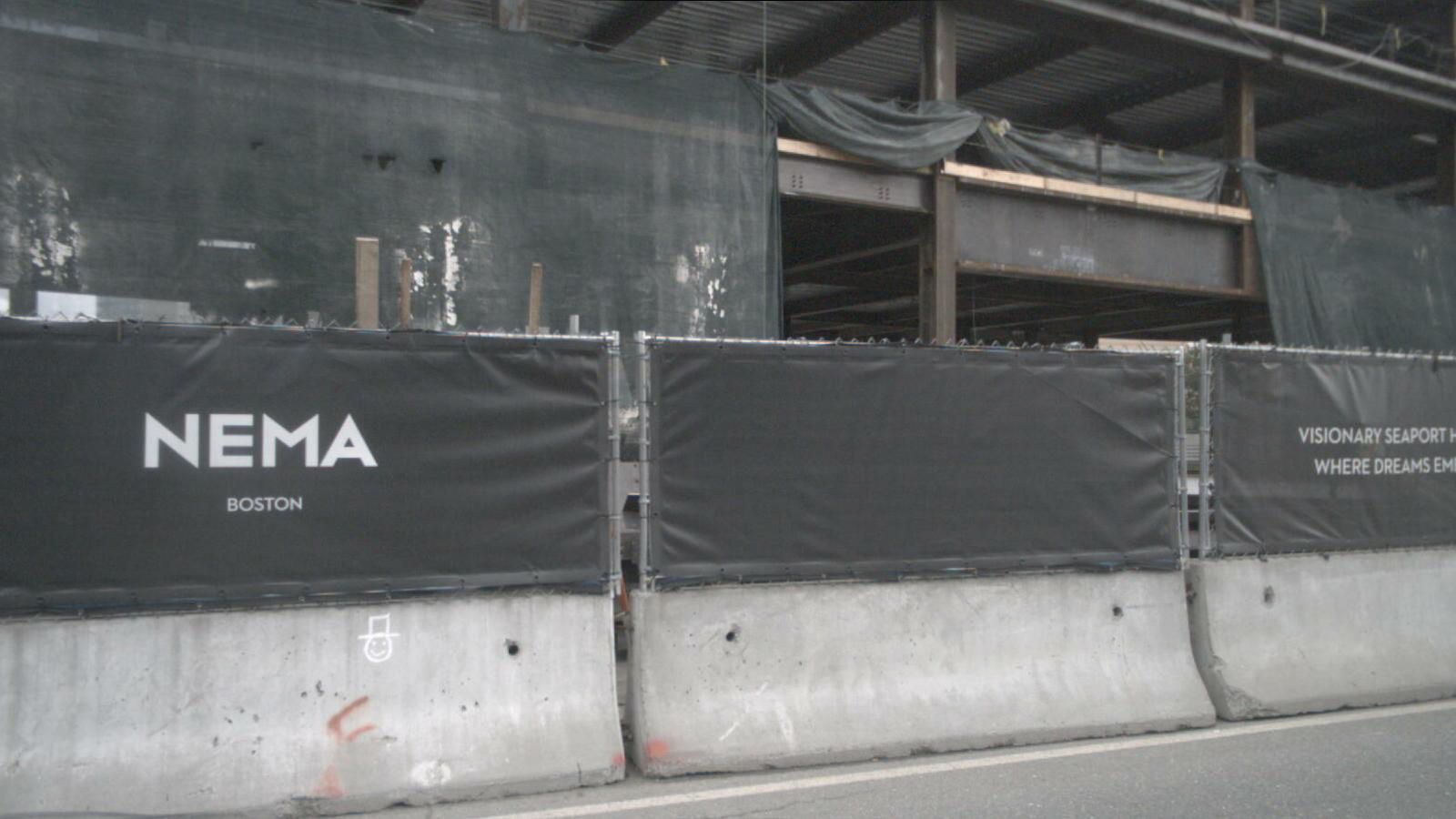}
\end{minipage}%
\hfill %
\begin{minipage}[b]{.74\textwidth}
  \centering
  \begin{subfigure}{.24\linewidth}
    \includegraphics[width=\linewidth]{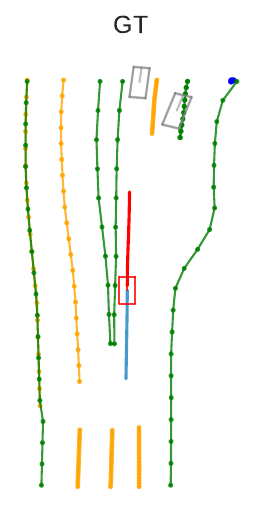}
    \label{fig:vis7_gt}
  \end{subfigure}%
  \begin{subfigure}{.24\linewidth}
    \includegraphics[width=\linewidth]{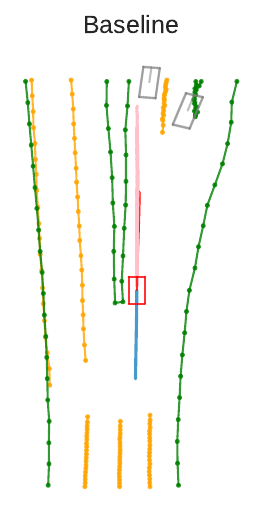}
    \label{fig:vis7_hmn}
  \end{subfigure}%
  \begin{subfigure}{.24\linewidth}
    \includegraphics[width=\linewidth]{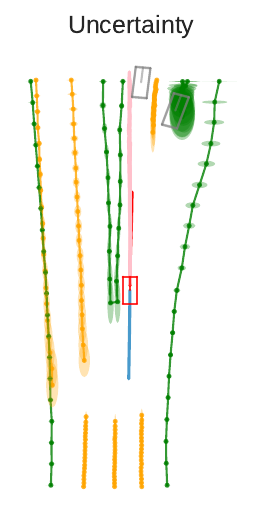}
    \label{fig:vis7_hmu}
  \end{subfigure}%
  \begin{subfigure}{.24\linewidth}
    \includegraphics[width=\linewidth]{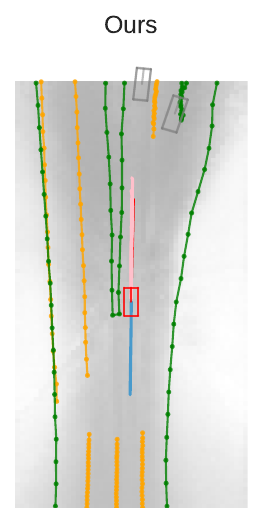}
    \label{fig:vis7_dsu}
  \end{subfigure}

    \vspace{-0.4cm}

    \begin{subfigure}{0.92\linewidth}
        \begin{tikzpicture}
        \draw [darkgreen, ultra thick] (0,0.5) -- (0.35,0.5) node[right, black] {\scriptsize Road Boundary};
        \draw [darkyellow, ultra thick] (2.65,0.5) -- (3,0.5) node[right, black] {\scriptsize Lane Divider};
        \draw [blue, ultra thick] (5.0,0.5) -- (5.35,0.5) node[right, black] {\scriptsize Pedestrian Crossing};
        \draw [red, ultra thick] (0,0) -- (0.35,0) node[right, black] {\scriptsize GT Future};
        \draw [pink, ultra thick] (2.1,0) -- (2.45,0) node[right, black] {\scriptsize Predicted Trajectories};
        \draw [historyblue, ultra thick] (5.65,0) -- (6,0) node[right, black] {\scriptsize Agent History};
        \end{tikzpicture}
  \end{subfigure}

\end{minipage}

\vspace{-0.2cm}

\caption{StreamMapNet~\cite{yuan2024streammapnet} and DenseTNT~\cite{GuSunEtAl2021} combined using the strategy in \cref{sec:strategy_3}. By replacing agent trajectory information with BEV features, DenseTNT is able to predict more accurate trajectories, compared to the significant overshooting outputs from the Baseline and Uncertainty-enhanced approaches.}
\label{fig:vis7}

\vspace{-0.6cm}

\end{figure*}

\label{sec:supp_vis}